%% file: low_rank_forecasting.tex
\documentclass[12pt, hidelinks]{article}

\usepackage[utf8]{inputenc}
\usepackage{amsmath, amssymb, url, float, graphicx, float,booktabs}
\usepackage{algorithm, algpseudocode}
\usepackage{fullpage}
\usepackage[small,bf]{caption}

\input defs.tex

\title{Low Rank Forecasting}
\author{Shane Barratt \and Yining Dong \and Stephen Boyd}
\date{\today}

\begin{document} 
\maketitle 
\begin{abstract}
We consider the problem of forecasting multiple values of the future of 
a vector time series, using some past values.
This problem, and related ones such as one-step-ahead prediction,
have a very long history, and there are a number of well-known methods
for it, including vector auto-regressive models, state-space methods,
multi-task regression, and others.  Our focus is on low rank forecasters,
which break forecasting up into two steps: estimating a vector that can 
be interpreted as a latent state, given the past, and then estimating the
future values of the time series, given the latent state estimate.
We introduce the concept of forecast consistency, which means that the estimates
of the same value made at different times are consistent.
We formulate the forecasting problem in general form, and focus on
linear forecasters, for which we propose a formulation that can be solved
via convex optimization.
We describe a number of extensions and variations,
including nonlinear forecasters, data weighting, the inclusion of auxiliary data,
and additional objective terms.
We illustrate our methods with several examples.
\end{abstract}

\section{Introduction}\label{s-intro}

\paragraph{Forecasting.}
We consider the problem of forecasting future values of a vector time 
series $x_t\in\reals^n$, $t=1,2,\ldots$, given previously observed values.
At each time $t$ we form an estimate of the future values $x_{t+1}, \ldots, x_{t+H}$,
where $H$ is our prediction horizon.  We denote these as $\hat x_{t+1|t}, \hat x_{t+2|t}, 
\ldots, \hat x_{t+H|t}$,
where $\hat x_{\tau|t}$ is our prediction of $x_\tau$ made at time $t$.
These estimates are based on the $M$ current and past values, 
$x_t, x_{t-1}, \ldots, x_{t-M+1}$, where $M$ is the memory of our forecaster.
When $H=1$, forecasting reduces to the common problem of predicting the next value
in the time series, given the previous $M$ values.

We introduce some notation to denote these windows of past and future values.
We define the \emph{past} at time $t$ as 
\[
p_t=(x_{t-M+1},\ldots,x_{t}) \in \reals^{Mn}, \quad t=M, M+1, \ldots .
\]
We define the \emph{future} at time $t$ as 
\[
f_t=(x_{t+1},\ldots,x_{t+H}) \in \reals^{Hn}, \quad t=1, 2, \ldots.
\]
We make the observation that $p_t$ and $p_{t+1}$ are related by a block shift, since
\[
p_{t+1}=(x_{t-M+2},\ldots,x_{t+1}) = ((p_t)_{n+1:Mn}, x_{t+1}),
\]
where $a_{i:j}$ denotes the subvector of $a$ with entries $i,\ldots, j$.
A similar shift structure holds for $f_t$.

We use $\hat f_t$ to denote our estimate or forecast of the future at time $t$,
\ie, 
\[
\hat f_t = (\hat x_{t+1|t},\ldots,\hat x_{t+H|t}) \in \reals^{Hn}.
\]
A \emph{forecaster} has the form $\hat f_t = \phi(p_t)$,
and $\phi:\reals^{Mn}\to\reals^{Hn}$ defines the forecaster.

We observe that the forecasts do not necessarily have the shift structure
that the past and future vectors do.
While $(f_{t+1})_{1:n}$ and $(f_{t})_{n+1:2n}$ are both equal to $x_{t+2}$,
$(\hat f_{t+1})_{1:n}$ and $(\hat f_{t})_{n+1:2n}$ can be different.
The first is $\hat x_{t+2|t+1}$, our estimate of $x_{t+2}$ made at time $t+1$,
whereas the second is
$\hat x_{t+2|t}$, our estimate of $x_{t+2}$ made at time $t$.  These two estimates need
not be the same.
(We will come back to this soon with the concept of forecasting consistency.)

\paragraph{Low rank forecasting and latent state.} We say that the forecaster has rank $r$ if
the forecast function factors as $\phi = \mathcal V \circ \mathcal U$, where
$\mathcal U: \reals^{Mn}\to \reals^r$ is the \emph{encoder}
and $\mathcal V: \reals^r \to \reals^{Hn}$ is the \emph{decoder}, and
$r \leq \min\{Mn,Hn\}$.
This means that the forecast takes place in two steps:
we first form the intermediate $r$-vector $z_t = \mathcal U(p_t)$ by encoding the
past, and then compute the forecast as $\hat f_t = \mathcal V(z_t)$ by decoding $z_t$. 
We can interpret the time series $z_t$, $t = M, M + 1, \ldots$ as a \emph{latent state}.
The term state is justified since $z_t$ is a summary of the past sufficient to carry out
our forecast.
Under Kalman's definition, the state of a dynamic system is
``the least amount of data one has to know about the
past behavior of the system in order to predict its future behavior'' \cite{kalman_new_1960}.
(We note one small difference:
In the traditional abstract definition of state, the past and
future are infinite, whereas here we have limited them to $M$ and $H$ time periods,
respectively.) 
The term rank for the dimension of the intermediate value $z_t$ is not standard;
but we will see later that it coincides with the rank of a certain matrix 
when we restrict our attention to linear forecasters.

The latent state can be very useful in applications, since it summarizes
what we need to know to about the past at time $t$ in one vector $z_t \in \reals^r$,
in order to carry out our forecast.
In some applications, identifying the latent state can be just as important
as carry out the actual forecasts.

\paragraph{Judging forecaster performance.}
Suppose we have a $T$-long observation of the time series, $x_1, \ldots, x_T$,
with $T \geq M+H$.
From this data set we extract the $N$ pairs of past and future,
\[
(p_M,f_M), \ldots, (p_{T-H}, f_{T-H}),
\]
with $N=T-H-M+1$.
We judge the performance of a forecaster on this data set 
by its average loss,
\[
\mathcal L = \frac{1}{N}\sum_{t=M}^{T-H} 
\ell(\hat f_t - f_t),
\]
where $\ell: \reals^{Hn} \to \reals$ is a convex loss function.
(The lower the loss function, the better the forecast.)
Common choices include the $\ell_2$ (squared) loss 
$\ell(u) = \|u\|_2^2$, $\ell_1$ loss $\ell(u)=\|u\|_1$,
or an appropriate Huber penalty function \cite[\S 6.1.2]{boyd_convex_2004}.

When the data set is also the one used to fit or choose the forecaster,
$\mathcal L$ is the training loss.  When the data set is a different set of 
data, not used to fit or choose the forecaster, $\mathcal L$ is the test loss.
We are interested in finding a forecaster that has low test loss, \ie,
makes good forecasts on data that was not used to fit it.
Another approach is walk-forward cross-validation,
where one produces a number of successive training and test data sets
from a single data set, where all test data points occur
after all training data points.
(This is to avoid look-ahead bias, and is in contrast to standard
cross-validation, where one creates random training and test splits.)

\paragraph{Forecaster consistency.}
Consider the value $x_\tau$, with $\tau \geq M+1$. 
We make predictions of $x_\tau$, denoted as $\hat x_{\tau|t}$,
at times
\[
t= \max\{\tau-H,M+1\}, \ldots, \min\{\tau-1,T-H\}.
\]
These 
\[
\min\{\tau,T-H+1\}-\max\{\tau-H,M+1\} = \left\{ \begin{array}{ll}
H & H+M+1\leq \tau \leq T-H+1,\\
\tau-M-1 & \tau \leq \min\{T-H+1,H+M+1\},\\
T-\tau+1 & \tau \geq \max\{T-H+1,H+M+1\},\\
T-H-M & \text{otherwise},
\end{array}
\right.
\]
forecasts of the same value,
made at different times and with different available information,
need not be the same.
We say the forecast is \emph{consistent} if these forecasts are 
not too different.
We note that inconsistency is not necessarily bad; it simply means that 
over the different periods
in which we form a prediction of $x_\tau$, we are changing our prediction.

While other measures of inconsistency could be used, we will judge inconsistency 
of the forecasts of $x_\tau$ using a sum of squares measure.
Define
\[
\bar x_\tau = \frac{1}{\min\{\tau,T-H+1\}-\max\{\tau-H,M+1\}} \sum_{t=\max\{\tau-H, M+1\}}^{\min\{\tau-1,T-H\}} \hat x_{\tau|t},
\]
the average of the predictions of $x_\tau$ made at different times.
Thus $\hat x_{\tau|t}-\bar x_\tau$ is the deviation of 
the prediction of $x_\tau$ made at time $t$ and the average of all
predictions we make of $x_\tau$.

Now consider a $T$-long observation of the time series, from which we obtain
the data set $(p_t,f_t)$, $t=M, \ldots, T-H$.
We define the (sum of squares) inconsistency as
\[
\mathcal I = \sum_{\tau= M+1}^{T} 
~\sum_{t=\max\{\tau-H,M+1\}}^{\min\{\tau-1,T-H\}} \| \hat x_{\tau|t} - \bar x_\tau \|_2^2.
\]

While forecaster consistency need not lead to better performance
(and indeed, often does not),
it can be considered as a desirable property for a forecaster, 
independent of forecast loss or performance \cite{hilary2013analyst}.
As a concrete example, suppose we are predicting the future cash flows of a business,
and adjusting business operations based on these forecasts.
In this case, inconsistent forecasts could lead to more changes in business operations
than we would like.  We may prefer forecasts that are more consistent,
at the cost of some loss in forecast performance.

\paragraph{Statistical forecasting.}
We mention here a general method for forecasting that includes many 
existing methods (described in more detail below).
Suppose we assume that $\{x_t\}$ is a stationary stochastic process,
with a distribution of $(p_t,f_t) \in \reals^{M+H}$
that, by stationarity, does not depend on $t$.
A natural forecast in this case is
the conditional mean of the future given the past, \ie,
\[
\hat f_t = \phi(p_t) = \Expect (f_t \mid p_t).
\]
(The forecaster function $\phi$ does not depend on $t$.)
This forecaster minimizes the mean square error $\Expect \|\hat f_t - f_t\|_2^2$
over all possible forecasters.

\paragraph{Outline.}
In \S\ref{s-linear-forecasting} we describe the special case of linear forecasting.
In \S\ref{s-linear-forecasting-methods} we describe a number of methods for 
producing linear forecasters.
In \S\ref{s-our-method} we describe our method for low rank forecasting.
In \S\ref{s-extensions-variations} we describe a number of extensions and variations,
a number of which have been incorporated in the software.
In \S\ref{s-examples} we show three examples: simulated, stock volatility,
and traffic.
We defer an extended discussion of the very large body of 
prior and related work to \S\ref{s-related-work}.

\section{Linear forecasting}
\label{s-linear-forecasting}

We now focus on linear forecasters, which have the form
$\phi(p_t) = \theta^T p_t$, where $\theta\in\reals^{Mn}\times\reals^{Hn}$ 
is the forecaster parameter matrix. 
We partition $\theta$ as $\theta=\begin{bmatrix}\theta_1 \cdots \theta_H\end{bmatrix}$,
so we have
\[
\hat x_{\tau |t} = \theta_{\tau-t}^T p_t, \quad \tau=t+1,\ldots,t+H.
\] 

\paragraph{Low rank linear forecasting.}
The general concept of low rank forecasting introduced in \S\ref{s-intro} 
is very simple in the case of a linear forecaster.  
Suppose $\phi= \mathcal V \circ \mathcal U$, with
$\mathcal V$ and $\mathcal U$ both linear, 
say $\mathcal U(p_t) = U^T p_t = z_t$ and
$\mathcal V(z_t) = V^T z_t$, with
$U\in\reals^{Mn\times r}$ and $V\in\reals^{r \times Hn}$.
Evidently we have $\theta = UV$, so the rank of $\theta$ is no more than $r$.
Conversely if $\theta$ has rank $r$, we can factor it to obtain $U$ and $V$.
So for a linear forecaster, a low rank coefficient matrix corresponds to a low
rank forecaster, in the general sense.

Evidently the latent state associated with a low rank linear forecaster
is only defined up to an invertible linear transformation, since $\tilde U =US$
and $\tilde V = S^{-1}V$ define the same forecaster, when 
$S \in \reals^{r\times r}$ is invertible.  The latent state associated
with $\tilde U$ and $\tilde V$ is $S^Tz_t$, where $z_t$ is the latent state
associated with $U$ and $V$.

\paragraph{Hankel data matrices.}
Suppose we have data $x_1, \ldots, x_T$,
from which we extract $N=T-H-M+1$ past/future pairs,
$(p_M,f_M), \ldots, (p_{T-H},f_{T-H})$.
From these data we form the data matrices
\[
P = \begin{bmatrix}
p_M^T \\
\vdots \\
p_{T-H}^T
\end{bmatrix}\in\reals^{N\times Mn}, \qquad
F = \begin{bmatrix}
f_M^T \\
\vdots \\
f_{T-H}^T
\end{bmatrix} \in\reals^{N \times Hn}.
\]
These matrices are block Hankel, due to the shift structure
mentioned in \S\ref{s-intro}.

\paragraph{Forecasts matrix.}
We stack the forecasts into the matrix
\[
\hat F = \begin{bmatrix}
\hat f_M^T \\
\vdots \\
\hat f_{T-H}^T
\end{bmatrix} = P\theta.
\]
This matrix is in general \emph{not} block Hankel, unless the forecaster is completely
consistent, \ie, never changes its prediction of any value $x_\tau$.
(This only happens when $\mathcal I=0$.)

\paragraph{Loss.}
The average loss can be expressed as
\[
\mathcal L = \frac{1}{N} \ones^T \ell(P\theta-F),
\]
where we extend $\ell$ to apply row-wise to its matrix argument, and $\ones$
is the vector with all entries one.  For the $\ell_2$ (squared) loss, this can be 
written as $\mathcal L = (1/N)\|P\theta-F\|_F^2$, where 
$\| \cdot \|_F$ denotes the Frobenius norm of a matrix.

\paragraph{Inconsistency.}
The inconsistency measure $\mathcal I$ can be expressed as
\[
\mathcal I = \dist (\hat F)^2,
\]
where $\dist (\hat F)$ is the Frobenius norm distance to the set of block Hankel
matrices.
It is readily shown that the projection $\pi(Z)$ of an $N\times Hn$ matrix $Z$ onto the 
set of block Hankel matrices is obtained by replacing each block by the average
over the corresponding anti-diagonal blocks.
We observe for future use that $\mathcal I$ is a convex quadratic function of $\theta$.
The inconsistency measure can be evaluated in $O(NHn)$ flops,
and its gradient can be evaluated in the same order; see 
appendix \S\ref{s-gradients} for the details.

\section{Linear forecasting methods}
\label{s-linear-forecasting-methods}

In this section we describe several general and well known methods for constructing
a linear forecaster.
Our purpose here is to describe these forecasting methods using
our notation; we will not use the material of this section in the sequel.
Some of the methods described here produce low rank forecasters.
In other cases, if a low rank forecaster is desired, we can use the truncated SVD 
(singular value decomposition) of the coefficient matrix
to obtain a low rank approximation.

\subsection{Forecasting using statistical models}
\label{sec:statistical_models}

\paragraph{Forecasting via autocovariance.}
The methods described below start by modeling $(p_t,f_t)$ as
a Gaussian zero mean random variable (that does not depend on $t$),
\[
(p_t, f_t) \sim \mathcal N(0, \Sigma).
\]
Since $(p_t,f_t) = (x_{t-M+1}, \ldots, x_{t+H})$, $\Sigma$ is block Toeplitz,
\[
\Sigma = \left[ \begin{array}{ccccc}
\Sigma_0 & \Sigma_1 &  \cdots  & \Sigma_{M+H-1} \\
\Sigma_{-1} & \Sigma_0 &  \cdots & \Sigma_{M+H-2} \\
\vdots & \vdots & \ddots &  \vdots  \\
\Sigma_{-M-H+1} & \Sigma_{-M-H+2} & \cdots & \Sigma_0  \\
\end{array}\right],
\]
where
$\Sigma_{i} = \Expect x_tx_{t+i}^T$ is the $i$th autocovariance
of the process $\{x_t\}$ \cite{gubner2006probability}.
(So $\Sigma_{-i}=\Sigma_i^T$.)
There are many methods available to estimate the 
autocovariance matrices
$\Sigma_0, \ldots, \Sigma_{M+H-1}$ from some training data;
see, \eg, \cite{box_time_2008,hastie2009elements,pourahmadi2013high}.

To obtain a forecaster, we partition $\Sigma$ as
\[
\Sigma = 
\begin{bmatrix}\Sigma_\text{pp} & \Sigma_\text{pf} \\ \Sigma_\text{fp} & \Sigma_\text{ff}\end{bmatrix},
\]
with $\Sigma_\text{pp} \in \reals^{Mn \times Mn}$ and $\Sigma_\text{ff}\in \reals^{Hn\times Hn}$.
Using a conditional mean forecaster, we have
\BEQ\label{e-gaussian-cond-mean}
\phi(p_t) = \Expect( f_t \mid p_t )= \Sigma_\text{fp} \Sigma_\text{pp}^{-1} p_t = \theta^T p_t,
\EEQ
where $\theta=\Sigma_\text{pp}^{-1}\Sigma_\text{pf}$.
This forecaster is not, in general, low rank.

\paragraph{Iterated AR($M$) forecasting.}
Another approach is to posit an AR($M$) model of $x_t$, \ie,
\BEQ
x_{t+1} = \sum_{i=1}^MA_i x_{t-i+1} + \epsilon_t, \quad t=M,M+1,\ldots,
\label{eq:ar}
\EEQ
where $\epsilon_t \sim \mathcal N(0,W)$ are independent
and $A_i\in\reals^{n \times n}$, $i=1,\ldots,m$.
We implicitly assume that the coefficients are such that the model is stable, so 
it defines a stationary stochastic process.
A simple way to fit an AR($M$) model is by linear regression \cite{burg1968new};
$W$ is an estimate of the one-step ahead prediction error covariance.

With this model we can work out the autocovariance matrices 
and then use the general formula \eqref{e-gaussian-cond-mean},
but we can more directly find the conditional mean of the future, given
the past.
Evidently we have 
\[
\Expect (x_{t+1} \mid p_t) = \sum_{i=1}^MA_i x_{t-i+1}
= \left[ \begin{array}{cccc} A_M & A_{M-1} & \cdots & A_1
\end{array} \right] p_t,
\]
and, continuing recursively, for $\tau=t+2, \ldots$, we have
\[
\Expect (x_\tau \mid p_t ) =
\sum_{i=1}^M A_i \tilde x_{\tau-i\mid t},
\]
where $\tilde x_{s \mid t} = x_t$ if $s\leq t$ and $\tilde x_{s \mid t} = \Expect (x_s \mid p_t)$ for
$s >t$.
This is the same as iterating the dynamics of the AR($M$) model forward, with 
$\epsilon_s=0$.
This forecaster is evidently linear, but not, in general, low rank.
(Of course it agrees with the general formula \eqref{e-gaussian-cond-mean} above.)

\paragraph{Forecasting using a latent state space model.}
Another approach is to model $\{x_t\}$ as a stationary Gaussian stochastic process generated by a state
space model \cite{gilbert1963controllability,kalman_new_1960,kalman1963mathematical,aoki2013state},
\[
z_{t+1} = Az_t + \epsilon_t, \quad
x_t= Cz_t + \eta_t, \quad t=1,2, \ldots,
\]
where $z_t\in\reals^r$ is the latent state,
$A\in\reals^{r \times r}$, and $C\in\reals^{n \times r}$,
$\epsilon_t\sim\mathcal N(0,Q)$ is a process noise,
and $\eta_t\sim\mathcal N(0,R)$ is a measurement noise.
(We assume that $\epsilon_t$ and $\eta_t$ are independent of one another
and across time.)
We assume that the matrix $A$ is stable (\ie, its eigenvalues have magnitude less
than one), so this defines a stationary stochastic process.
There are many ways to fit a state space model to time series training data,
\eg, via N4SID \cite{van_overschee_n4sid_1994}, EM \cite{shumway_approach_1982},
and least squares auto-tuning \cite{barratt_fitting_2020}.
Here too we can work out the autocovariance coefficients and use the general
formula \eqref{e-gaussian-cond-mean} above to find a forecaster.  
This forecaster, not surprisingly, has low rank, indeed, rank $r$.

Here we describe the construction of the associated forecaster 
in a more natural way, that explains the factorization of $\theta$ into
two natural parts: a state estimator, followed by a forward simulator.
The first step is to determine $\Expect(z_t \mid p_t)$, our estimate of 
the current latent state, given the past.
This can be done by solving the Kalman smoothing problem \cite{kalman_new_1960} with memory $M$
\[
\begin{array}{ll}
\text{minimize} & \sum_{i=1}^{M-1} \epsilon_{t-i}^TQ^{-1}\epsilon_{t-i} + \sum_{i=0}^{M-1} \eta_{t-i} R^{-1} \eta_{t-i} \\
\text{subject to} & \epsilon_\tau = z_{\tau+1} - Az_\tau, \quad \tau= t-M+1,\ldots,t-1, \\
& \eta_\tau = x_\tau - Cz_\tau, \quad \tau=t-M+1,\ldots,t,
\end{array}
\]
with variables 
\[
z_s, \quad \epsilon_s, \quad  \eta_s, \quad s=t-M+1, \ldots, t.
\]
The value of $z_t$ is $\Expect(z_t \mid p_t)$.
This is a linearly constrained least squares problem,
so the solution is a linear function of the past $p_t$, \ie,
$z_t = K p_t$, where $K\in\reals^{r \times Mn}$.
The matrix $K$ is readily found by forming the KKT optimality conditions,
and standard linear algebra computations \cite{boyd_convex_2004,boyd_introduction_2018}.
We note one slight difference between this state estimator and the traditional
Kalman filter: this one uses only the $M$ previous values of $x_s$ (\ie, $p_t$),
whereas the Kalman filter uses all previous values.

It is readily seen that for $\tau = t+1, \ldots, t+H$, we have 
\[
\Expect (z_\tau \mid p_t) = A^{\tau-t} \Expect (z_t \mid p_t) = 
A^{\tau-t} Kp_t,
\]
and 
\[
\Expect( x_\tau \mid p_t) = C\Expect (z_\tau \mid p_t) =
C A^{\tau-t} Kp_t.
\]

So we have
\[
\hat f_t = \begin{bmatrix} CA \\ CA^2 \\ \vdots \\ CA^H\end{bmatrix} Kp_t.
\]
This shows that $\theta$ has rank (at most) $r$, 
the dimension of the latent time series.
This is hardly surprising, since under this model the past and future are independent,
given the current state $z_t$.

\subsection{Forecasting via regression}

Regression yields another set of methods for choosing $\theta$ directly
from a given training data set.
(Regression methods and statistical methods are closely linked,
as we discuss below.)
Let $\mathcal L(\theta)$ denote the loss on the training data set.
In (regularized) regression, the parameter matrix $\theta$ is chosen as a minimizer of
\[
\mathcal L(\theta) + \lambda \mathcal R(\theta),
\]
where $\mathcal R: \reals^{Mn \times Hn}\to \reals$ is a convex 
regularizer function, and $\lambda$ is a positive hyper-parameter.
This is a convex function, so computing an optimal $\theta$ is 
in principle straightforward.

The most common regression problem, ridge regression,
uses $\ell_2$ loss and regularization, so the objective is
\BEQ
\frac{1}{N}\|P\theta-F\|_F^2 + \lambda \|\theta\|_F^2.
\label{eq:sep_reg}
\EEQ
with variable $\theta$.
This has the minimizer
\[
\theta = (P^TP + N \lambda I)^{-1} P^TF.
\]
The objective is separable across the columns of $\theta$, which means each column
of $\theta$ can be found separately.
This forecaster simply uses a separate ridge regression to 
predict $x_\tau$ based on $p_t$, for $\tau=t+1, \ldots, t+H$.

When $\lambda =0$, ridge regression is ordinary least squares, and coincides
with the statistical method above when we use the empirical estimates
of the autocovariance matrices,
since $\theta = \hat \Sigma_\text{pp}^{-1} \hat \Sigma_\text{fp}$, where
$\hat \Sigma_\text{pp}$ is the empirical covariance of $p_t$
and $\hat \Sigma_\text{fp}$ is the empirical covariance of $f_t$ and $p_t$,
which are given by
\[
\hat \Sigma_\text{pp} = \frac{1}{N}P^TP, \qquad \hat \Sigma_\text{fp} = \frac{1}{N}P^TF.
\]

\section{A regularized regression method} \label{s-our-method}
The method we propose is a regression method
with two particular regularization functions,
one that encourages $\theta$ to be low rank,
and another that encourages forecaster consistency.
We choose $\theta$ to minimize
\BEQ
\mathcal L(\theta)
+  \lambda \|\theta\|_* + \kappa \mathcal I(\theta),
\label{eq:main_prob}
\EEQ
where $\|\cdot \|_*$ is the dual norm of a matrix
(also known as the nuclear norm, trace norm, Ky Fan norm, or Schatten norm),
\ie, the sum of its singular values,
and $\lambda$ and $\kappa$ are positive hyper-parameters
that control the strength of the two types of regularization.
The objective \eqref{eq:main_prob} is a convex function of $\theta$,
and so in principle straightforward to minimize.
The nuclear norm is widely used as a convex surrogate for the 
(nonconvex) rank function; roughly speaking it promotes low rank of its
matrix argument.  Generally (but not always),
the larger $\lambda$ is, the lower the rank of $\theta$.
The forecaster consistency term, as discussed above,
encourages $\theta$ to produce consistent forecasts.

\paragraph{Critical value of $\lambda$.}
There is a critical value $\lambda^\mathrm{max}$, with the property that
$\theta=0$ if and only if $\lambda \geq \lambda^\mathrm{max}$.
When $\mathcal L$ is differentiable, $\lambda^\mathrm{max}$ is given by
\[
\lambda^\mathrm{max} = \|\nabla_\theta \mathcal L(0)\|_2,
\]
where $\nabla_\theta \mathcal L(0)\in\reals^{Mn \times Hn}$ is the gradient
of $\mathcal L(\theta)$ at $\theta=0$,
and $\|\cdot\|_2$ is the $\ell_2$ norm (maximum singular value).
This can be verified by examining the condition under which 
$\theta=0$ is optimal, \ie, that the
subdifferential of \eqref{eq:main_prob} includes $0$.
When the $\ell_2$ loss is used, this condition reduces to
\[
\lambda^\mathrm{max} = \frac{2}{N}\|P^TF\|_2.
\]
(This can be computed efficiently, without forming the matrix $P^TF$,
using power iteration.)
It is convenient to express the nuclear norm regularization as
$\lambda = \alpha \lambda^\mathrm{max}$, with $\alpha \in [0,1]$.

\paragraph{Choosing $\alpha$ and $\kappa$.}
The traditional method for choosing the hyper-parameters $\alpha$ and
$\kappa$ is to compute $\theta$ for a number of combinations of them,
and for each forecaster, evaluate the performance on another (test) data set,
not used to form $\theta$, \ie, train the forecaster.
Among these forecasters we choose one that yields least or nearly 
least test loss, skewing toward larger values of $\alpha$ and $\kappa$.

When consistency is regarded as a second objective, and not a regularizer
meant to give better test performance, we fix $\kappa$
and do not consider it a hyper-parameter.  In this case the 
term $\kappa \mathcal I(\theta)$ should also be
included when evaluating the test performance.

\subsection{Solution method}

The objective \eqref{eq:main_prob} is convex, and can be minimized
using many methods.  Smaller instances of the problem can be solved
with just a few lines of generic CVXPY code \cite{diamond_cvxpy_2016,agrawal_rewriting_2018}.
There also exist a number of specialized methods for problems with nuclear
norm regularization \cite{ji_accelerated_2009,liu_nuclear_2014,toh_accelerated_nodate}.
Generic methods, however, will not scale well, since the number of 
scalar variables, $HMn^2$, can be very large when one or more of 
$H$, $M$, or $n$ is large.
We describe here a simple customized method that does scale well.
The method is closely related to well known methods
\cite{keshavan2012efficient,chen2012matrix,jain2013low,
hardt2013provable,netrapalli2014non,agarwal2016learning},
so we simply outline it here.

\paragraph{Factored problem.}
Suppose we know that the solution to \eqref{eq:main_prob}
has at most rank $k$.
Then \eqref{eq:main_prob} is equivalent to the \emph{factored} problem
\BEQ
\begin{array}{ll}
\text{minimize} & \frac{1}{N} \ones^T \ell(PUV - F) +
\lambda/2 (\|U\|_F^2 + \|V\|_F^2) + \kappa \mathcal I(UV),
\end{array}
\label{eq:alternating}
\EEQ
with variables $U\in\reals^{Mn\times k}$ and $V\in\reals^{k \times Hn}$.
That is, $U,V$ is a solution to \eqref{eq:alternating} if and only if
$\theta=UV$ is a solution to \eqref{eq:main_prob} and $\|U\|_F^2=\|V\|_F^2=\frac{1}{2}\|\theta\|_*$.
(See, \eg, \cite{udell_generalized_2016} for a proof.)
Solving this problem directly
gives us the two factors of $\theta$.
In fact we obtain a balanced factorization, \ie, 
the (positive) singular values of $U$ and $V$ are the same.

\paragraph{Alternating method.}
The problem \eqref{eq:alternating} is not convex, but it is convex
in $U$ for fixed $V$ and convex in $V$ for fixed $U$.
Each of these minimizations can be carried out efficiently using, \eg,
the limited-memory Broyden Fletcher Goldfarb and Shanno (L-BFGS) method \cite{liu_limited_1989}.
This method requires storing a modest number of matrices with the same same 
sizes as $U$ and $V$,
and in each iteration, the computation of the gradient of the objective
with respect to $U$ or $V$.
(The gradients of the objective in \eqref{eq:alternating} are given in
appendix \S\ref{s-gradients}; in our implementation they are mostly computed
automatically using automatic differentiation techniques.)

\paragraph{Computing the rank of $UV$.}
After solving problem \eqref{eq:alternating},
it can be the case that the rank of $UV$ is less than $k$.
We can both find the rank of $UV$, compute reduced versions of $U$ and $V$,
and compute the reduced-rank SVD of $\theta$ efficiently (\ie, without actually computing
the full SVD of $\theta=UV$ or even forming $\theta$) as follows.
Denote the rank of $U$ by $r_U$ and the rank of $V$ by $r_V$.
First, we compute the SVD of $U$ and $V$,
\[
U = U_U \Sigma_U V_U^T, \quad V = U_V \Sigma_V V_V^T,
\]
where $U_U,\Sigma_U,V_U$ and $U_V,\Sigma_V,V_V$ are the appropriate sizes.
Next we compute the SVD of the matrix $A = \Sigma_U V_U^T U_V \Sigma_V$,
which has rank $r=\rank(UV)$,
\[
A = U_A \Sigma_A V_A^T,
\]
where $U_A,\Sigma_A,V_A$ are the appropriate sizes.
Then the reduced rank versions of $U$ and $V$, denoted $\tilde U\in\reals^{Mn \times r}$ and $\tilde V\in\reals^{r \times Hn}$, are
given by
\[
\tilde U = U_U U_A \Sigma_A^{1/2}, \quad \tilde V = \Sigma_A^{1/2} V_A^T V_V^T.
\]
The SVD of $\theta$, $\theta=U_\theta \Sigma_\theta V_\theta$, is given by
\BEQ\label{e-theta-svd}
U_\theta = U_U U_A, \quad \Sigma_\theta = \Sigma_A, \quad V_\theta = V_VV_A.
\EEQ

\paragraph{Choosing $k$.}
We propose the following simple method for choosing the value of $k$.
We start with a modest value of $k$ (say 10 or 20),
solve \eqref{eq:alternating}, and then if $r=\rank(UV) =k$ (computed using the method
above), we double $k$ and solve \eqref{eq:alternating} again.
If, on the other hand, $r < k$, then we know that our choice of $k$
was large enough, and terminate.

\paragraph{Convergence.}
Since this is an alternating method, 
the objective is decreasing and so converges.
Whether the alternating method converges to the solution 
of the original (convex) problem is another question.
We can check for global optimality of $\theta=UV$ in the original 
convex problem as follows.
Suppose $\ell$ is differentiable and let
\[
G=\nabla_\theta \left(\frac{1}{N}\ones^T\ell(P\theta-F) + \kappa \mathcal I(\theta)\right),
\]
\ie, the gradient of the differentiable part of the objective.
Then $\theta$ is globally optimal if and only if
the following conditions hold
\[
\begin{aligned}
\|G + \lambda U_\theta V_\theta^T\|_2 &\leq& \lambda, \\
U_\theta^T G + \lambda V_\theta^T &=& 0,\\
G + \lambda V_\theta &=& 0,
\end{aligned}
\]
where $\theta=U_\theta \Sigma_\theta V_\theta$ is the SVD of $\theta$
(computed as described in \eqref{e-theta-svd} above).
The residuals of these three conditions could be used as a stopping 
criterion for the alternating method.

We have observed that in all numerical examples 
when $r<k$, the final
$\theta$ is close to satisfying the optimality conditions above.
Unfortunately, verifying optimality requires us to form a matrix the 
same size as $\theta$, as well as compute its norm.
While the norm could be evaluated using a power method, never explicitly
forming the matrix, we would suggest that this final global optimality 
check is not needed in practice.

\paragraph{Practical considerations.}
When we are solving the problem for many values of $\alpha$ and $\kappa$,
or performing walk-forward cross-validation,
we can warm-start this iterative algorithm at the previously computed solution.
It is worth noting that we do not need to actually form $P$ or $F$.
This can be necessary when the size of the original time series
fits in memory but $P$ and $F$ do not,
which could be the case when $M$ or $H$ is very large.
All we need is to compute the gradient of the objective with respect to $U$ or $V$,
which can be done without forming $P$ or $F$.
For example, $PU$ can be implemented as a one-dimensional
convolution of the time series $x_1,\ldots,x_T$ with a number of kernels extracted from $U$.
Since the alternating method only requires computing the gradient, and basic dense linear algebra,
it can be performed on either a CPU or GPU.

\section{Extensions and variations}
\label{s-extensions-variations}

In this section we describe a number of extensions and variations.
Several of these extensions are quite useful and have been incorporated into
the software.

\subsection{Nonlinear forecasting}

In this section we describe low rank nonlinear forecasting.
Our forecaster has the familiar form $\phi = \mathcal V \circ \mathcal U$,
but instead of the encoder and encoder being linear, they are
nonlinear functions, for example neural networks.
The encoder $\mathcal U$ encodes the past into the latent state $z_t$, as $z_t = \mathcal U(p_t;\theta_U)$
and has parameters $\theta_U\in\reals^{p_U}$.
The decoder $\mathcal V$ decodes
the state into the forecast $\hat f_t$, as $\hat f_t = \mathcal V(z_t;\theta_V)$
and has parameters $\theta_V\in\reals^{p_V}$.

The fitting problem in the nonlinear forecasting case becomes
\BEQ
\begin{array}{ll}
\text{minimize} & \frac{1}{N} \sum_{i=1}^N \ell(\hat f_t - f_t) +
\lambda/2 (\|\theta_U\|_F^2 + \|\theta_V\|_F^2) + \kappa \dist(\hat F)^2,\\
\text{subject to} & \hat f_t = \mathcal V(\mathcal U(p_t;\theta_U);\theta_V), \quad t=M,\ldots,T-H. 
\end{array}
\label{eq:non_lin}
\EEQ
The second term in the objective is no longer the nuclear norm of the forecaster matrix, since
the predictor is nonlinear.
However, it can still be useful; it will help control the complexity of the neural network parameters.
We can also use the same forecaster consistency term.

We can approximately solve problem \eqref{eq:non_lin} using the stochastic gradient method
(SGD).
We refer the reader to \cite{goodfellow_deep_2016} and the references therein
for possible architectures and training methods.
We note that problem \eqref{eq:non_lin} is equivalent to the methods described in \S\ref{s-our-method}
with single layer (\ie, linear) neural networks for $\mathcal U$ and $\mathcal V$.

\subsection{Data weighting}
We can weight the components of the loss function,
based on how much we care about particular parts of the forecast.
That is, we adjust the loss term to
\[
\frac{1}{N}\ones^T\ell(W\circ \hat F),
\]
where $W\in\reals_+^{N \times Hn}$
is the \emph{weight matrix},
and $\circ$ denotes the Hadamard or elementwise product.
We denote the (block) elements of $W$ as $w_{\tau \mid t}$, \ie,
\[
W = \begin{bmatrix} w_{M+1 \mid M}^T &  \cdots & w_{M+H \mid M}^T \\
\vdots & \ddots & \vdots \\
w_{T-H+1 \mid T-H}^T & \cdots & w_{T \mid T-H}^T
\end{bmatrix}
\]
The larger $(w_{\tau \mid t})_i$ is, the more we care about
forecasting the $i$th element of $x_\tau$ at time $t$.

There are many ways to construct a weight matrix.
One way is via exponentially decaying weighting on $t$,
$\tau$, and a separate constant weight for each element of the time series.
That is, we specify a halflife for $t$, denoted $h^t > 0$
and a halflife for $\tau$, denoted $h^\tau > 0$.
Then let the weights for time and forecast time be 
\[
w^t = \exp(\log(0.5) / h^t)^{\tau - t}, \quad w^\tau = \exp(\log(0.5) / h^\tau)^{T-\tau}.
\]
We also specify a weight for each element of the time series,
denoted $w^\mathrm{col}\in\reals_+^n$.
For example, if $(w^\mathrm{col})_i=0$, then we do not care about forecasting
$(x_t)_i$.
(But note that we do use $(x_t)_i$ to forecast the other elements of $x_t$.)
The weights are then given by the product of these three weights,
\[
w_{\tau \mid t} = w^t w^\tau w^\mathrm{col}.
\]

\subsection{Auxiliary data}
\label{s-auxiliary-data}
Often we have auxiliary information or data separate from the time series
that can be useful for forecasting future values of the time series.
Common examples are time-based features, such as hour, day of week, or month.  
These features could be useful for forecasting, but are clearly not worth
forecasting themselves.
Another example is an additional 
time series that is related to or correlated with our time series.

We denote the auxiliary information known at time $t$ by the 
vector $a_t\in\reals^p$.
There are two ways to incorporate auxiliary information into our forecasting
problem.
The first is to remove the effect of $a_t$ on $x_t$,
and then forecast the residual time series.
We might do this by solving the problem
\BEQ
\begin{array}{ll}
\text{minimize} & \sum_{t=1}^T \|Sa_t - x_t\|_2^2
\end{array}
\label{eq:detrending}
\EEQ
with variable $S\in\reals^{n \times p}$.
(We can of course add regularization here if needed.)
When the auxiliary information are simple functions of time
such as linear or sinusoidal,
this step is called de-trending or removing the trend or seasonality
from a time series 
(see, \eg, \cite[\S9]{box_time_2008}, \cite[\S13.1.1]{boyd_introduction_2018}, or
\cite[Appendix A]{velasquez-bermudez_dynamic_2019}).
We then define a new series $\tilde x_t = x_t - Sa_t$,
and forecast that series instead of the original series.
Our final forecast is then 
\[
\hat x_{\tau \mid t} = Sa_\tau+\hat{\tilde x}_{\tau \mid t},
\]
where $\hat{\tilde x}_{\tau \mid t}$
is the forecast for $\tilde x_\tau$ made at time $t$.
The first term is the baseline; the second is the forecast
of the residual time series, with the baseline removed.

The second way to incorporate auxiliary information 
is to make our forecaster a function of both the past
and auxiliary data.
That is, we let our forecaster be
\[
\hat f_t = \phi(p_t, a_t) = \theta^Tp_t + \Phi^T a_t,
\]
where $\Phi\in\reals^{p \times Hn}$.
We can decide whether or not to make $\Phi$ low rank.
If we want it to be low rank, then the nuclear norm regularization term
becomes
\[
\left\|
\begin{bmatrix}
\theta \\ \Phi
\end{bmatrix}
\right\|_*.
\]

\subsection{Other regularization}
Other convex regularization on $\theta$ can be added to problem \eqref{eq:alternating}
and the alternating method will work the same, because it will be biconvex in $U$ and $V$.
Convex regularization on the factors $U$ and $V$ can be added, and the alternating
method will work the same.
For example, if we wanted the encoder $U$ to be sparse, \ie, each element of the state
only depends on a few elements of the time series, then
we could add a multiple of 
the term $\sum_{i,j}|U_{ij}|$ to the objective in \eqref{eq:alternating}.
(We note however that L-BFGS does not handle nonsmooth terms like absolute value well,
so an alternate solution method might be needed.)
When regularization is added to $U$ or $V$ individually, the fitting problem is nonconvex.
While the algorithm will still work, there is no guarantee that it will converge to the global solution.

\subsection{Latent state dynamics}
If the sole objective is to forecast the original series $x_t$,
the latent series $z_t$ is simply an intermediate quantity used in forming
$\hat f_t$.
In other cases, the latent series $z_t$ discovered by our forecasting method
is actually of interest by itself.
In those cases, it might be interesting to look at its dynamics, \ie, how
$z_t$ evolves over time.
One reasonable model of $z_t$ is an autoregressive model,
\[
z_{t+1} = Az_t + \epsilon_t, \quad t=1,\ldots,T-1,
\]
where $A\in\reals^{r \times r}$ and $\epsilon_t\in\mathcal N(0, W)$ are independent.
We can fit such a model by linear regression.

\section{Examples}
\label{s-examples}
In this section we apply our method to three examples.
All experiments were conducted using PyTorch \cite{paszke_pytorch_2019} on an unloaded Nvidia 1080 TI GPU.

\subsection{Simulated state space dataset}
We consider a dataset sampled from a state space model,
\[
z_{t+1} = A z_t + \epsilon_t, \quad x_t = Cz_t + \eta_t,
\]
where the state is $z_t\in\reals^2$ and the observations are $x_t\in\reals^{10}$.
The entries of $A$ and $C$ are randomly sampled according to
\[
A_{ij} \sim \begin{cases}
\mathcal N(1, (0.1)^2) & i = j,\\
\mathcal N(0, (0.1)^2) & \text{otherwise},
\end{cases}
\qquad
C_{ij} \sim \mathcal N(0, 1).
\]
We scale $A$ so that its spectral radius is 0.98.
We set the covariance matrices to $Q=I$ and $R=(0.1)I$.
We consider as our training dataset a length 100 sample from the model,
and as our test dataset a length 500 sample from the model.
(In both of these datasets $z_1$ was sampled from the steady state distribution.)
We take $H=M=12$, so $\theta\in\reals^{120 \times 120}$
and we are using the 12 most recent values of $\{x_t\}$ to predict
the next 12 values of $\{x_t\}$.
The forecaster matrix $\theta$ contains $14400$ entries.
We use $\ell_2$ loss.

We begin by constructing the optimal (conditional mean) forecaster
using the techniques described in \S\ref{s-linear-forecasting}, and the 
actual (true) values of $A$, $C$, $Q$, and $R$.
This forecaster has a test loss of 10.54.  Aside from the small 
difference between expectation and the empirical loss over the test set,
no forecaster can do better,
since this forecaster minimizes mean square loss over all forecasters,
and uses the true values of the autocovariance matrices.
We can therefore consider $10.54$ as an approximate
lower bound on achievable performance.

\begin{figure}
\centering
\includegraphics[]{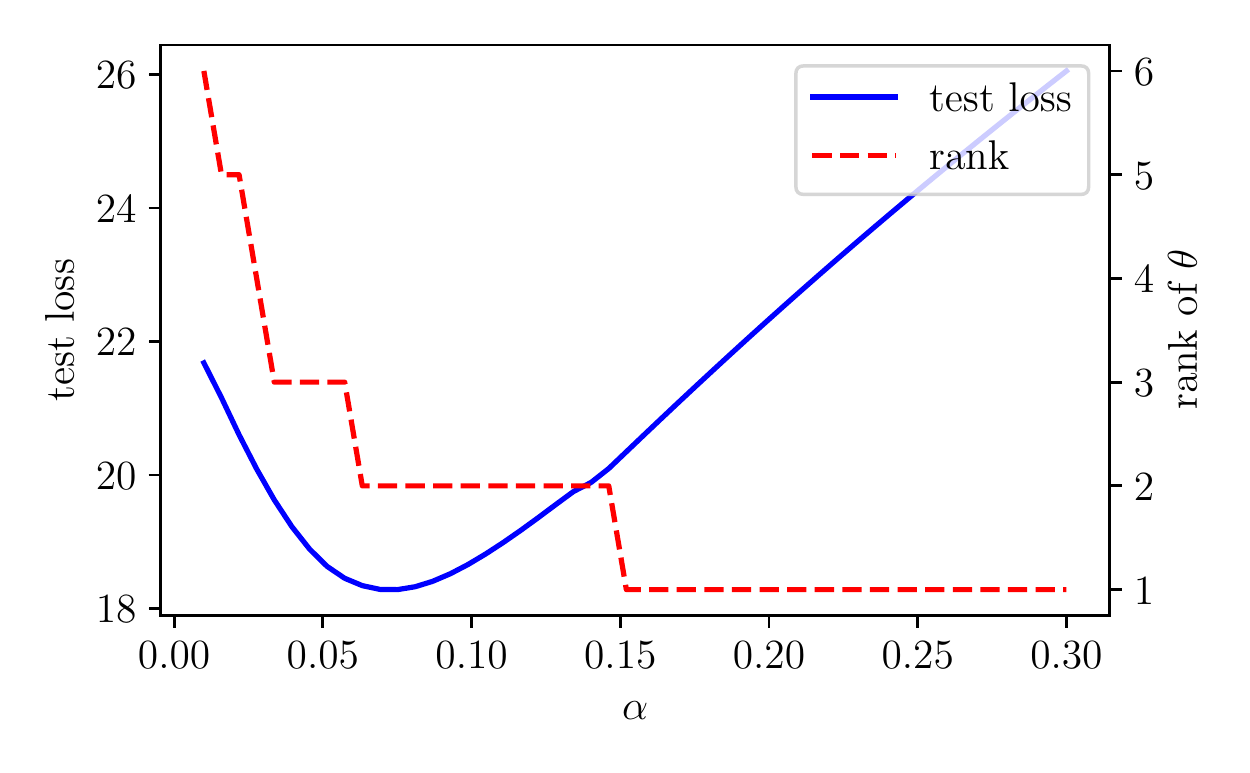}
\caption{Test loss and rank of $\theta$ versus $\alpha$ on the simulated dataset.}
\label{fig:simulated_test_loss}
\end{figure}

\begin{figure}
\centering
\includegraphics[]{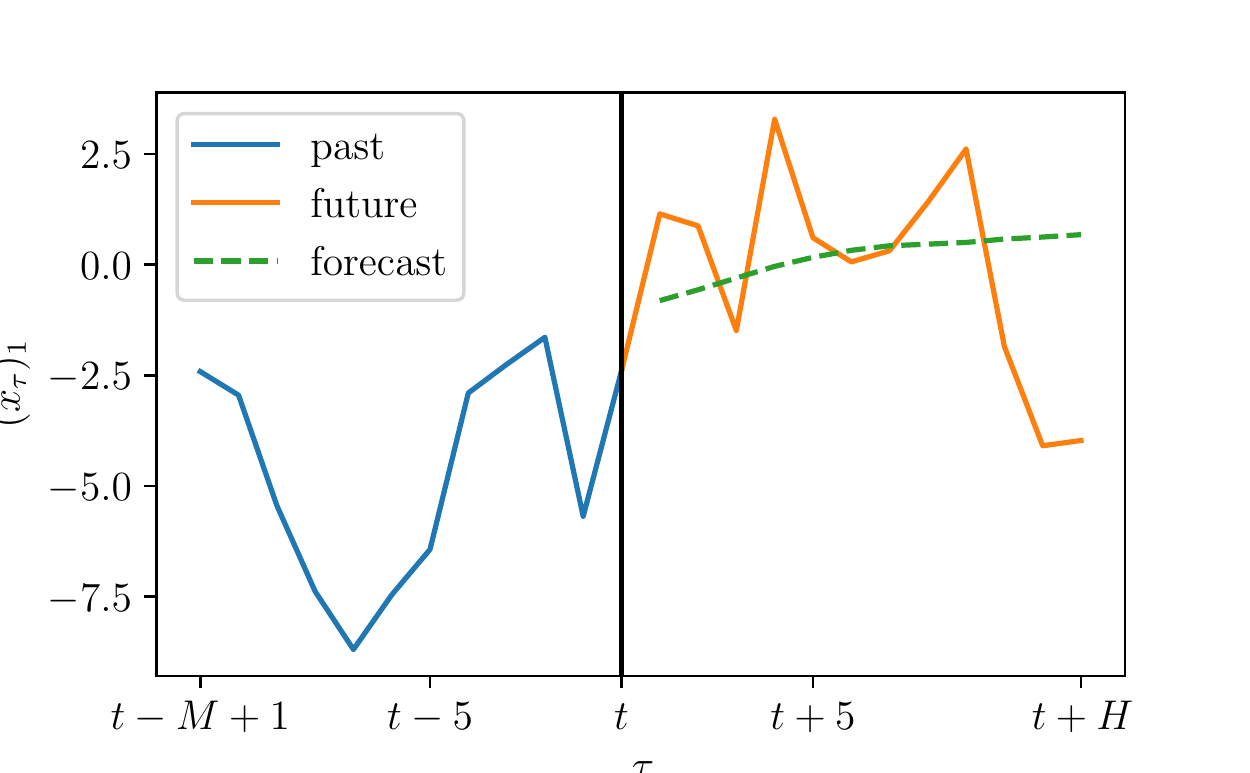}
\caption{Forecast of simulated data made at a time $t$, shown as vertical line.
The solid curve shows $(x_\tau)_1$, and the dashed
line is our forecast, $(\hat x_{\tau\mid t})_1$, $\tau=t+1,\ldots,t+H$.}
\label{fig:simulated_forecast}
\end{figure}

Next we apply our method to the training dataset
over 50 values of $\alpha\in[0.01,0.3]$,
with $\kappa=0$, and in figure \ref{fig:simulated_test_loss}
plot the test loss and rank of $\theta$ versus $\alpha$.
Fitting the forecaster took roughly five seconds.
(Warm starting the optimization from the previous value of $\alpha$ reduced this
considerably.)
As $\alpha$ increases, the rank goes down.
As $\alpha$ increases, the test loss initially goes down, and then
after a certain value, the test loss begins to go up.
This suggests that a good value of $\alpha$ is around 0.1, 
which corresponds to a forecaster of rank 2, which is the 
true dimension of the latent state.
The test loss of this forecaster is 18.28, a bit above the lower 
bound found when the exact values of $A$, $C$, $Q$, and $R$ are used.
For comparison, the test loss for the zero forecaster is $50.7$,
and the test loss for the empirical autocovariance forecaster is $27.23$.
In figure \ref{fig:simulated_forecast} we show a forecast
from our model (with $\alpha=0.1$) on the test dataset.

\begin{figure}
\centering
\includegraphics[]{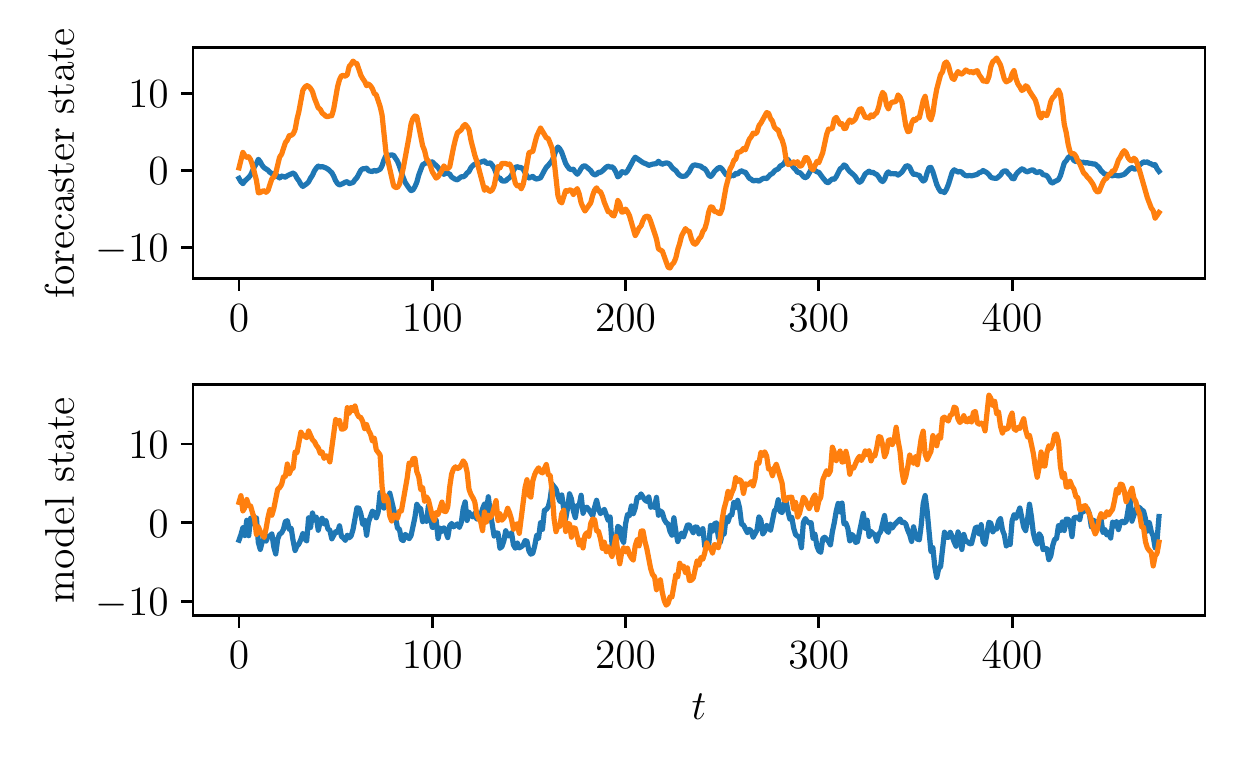}
\caption{Simulated dataset. \emph{Top.} Transformed latent state of the 
forecaster. \emph{Bottom.} True latent state of underlying
state space model.}
\label{fig:simulated_states}
\end{figure}

We can also compare our extracted latent state to the true latent state.
As mentioned above, the latent state is modulo a linear change of coordinates, 
so to compare the true latent state and the latent state of our forecaster,
we use our latent state to predict the true latent state
of the underlying state space model, by choosing $S$ to minimize
\[
\sum_{t=M}^{T-H}\|SU^Tp_t - z_t\|_2^2
\]
over the coordinate transformation $S\in\reals^{2 \times 2}$.
In figure \ref{fig:simulated_states} we plot the transformed states.
We can see that our transformed latent states reasonably track the 
true latent states.

\begin{figure}
\centering
\includegraphics[]{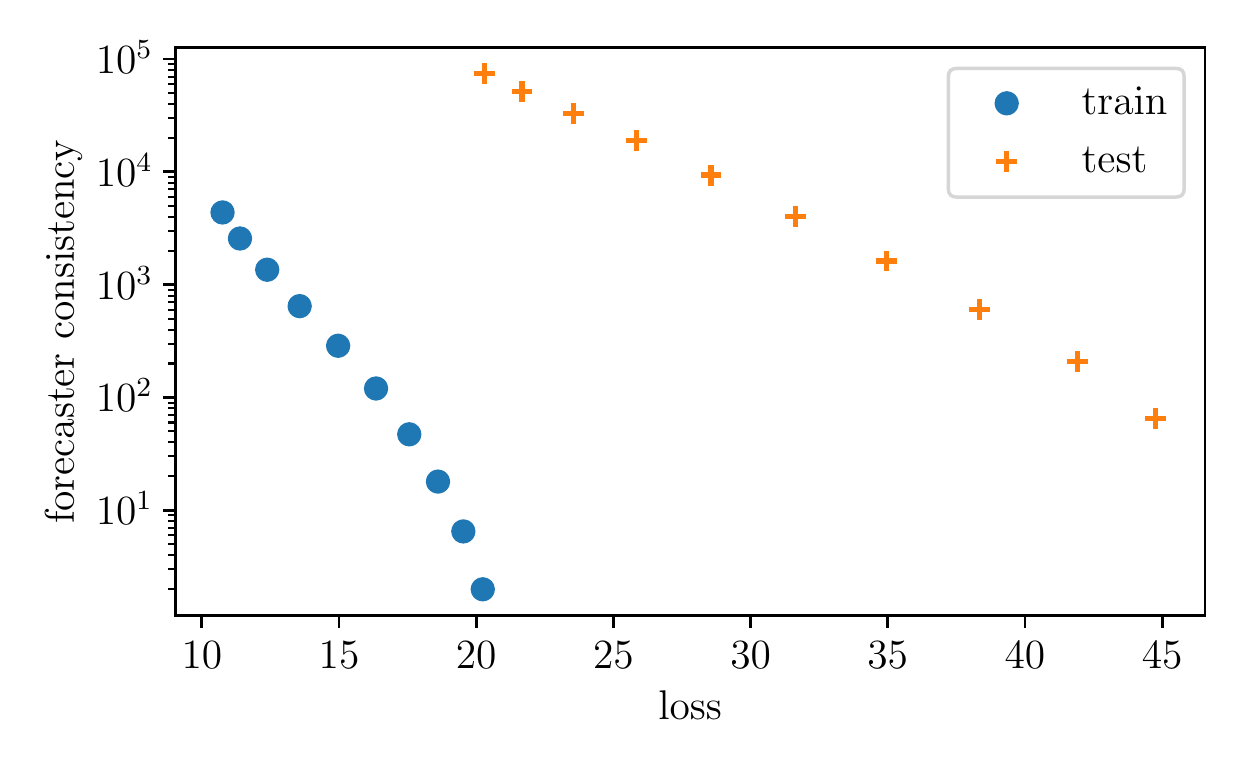}
\caption{Simulated dataset. Loss versus forecaster consistency on the train
and test dataset for a number of $\kappa$s.}
\label{fig:simulated_consistency_vs_test_loss}
\end{figure}

\begin{figure}
\centering
\includegraphics[]{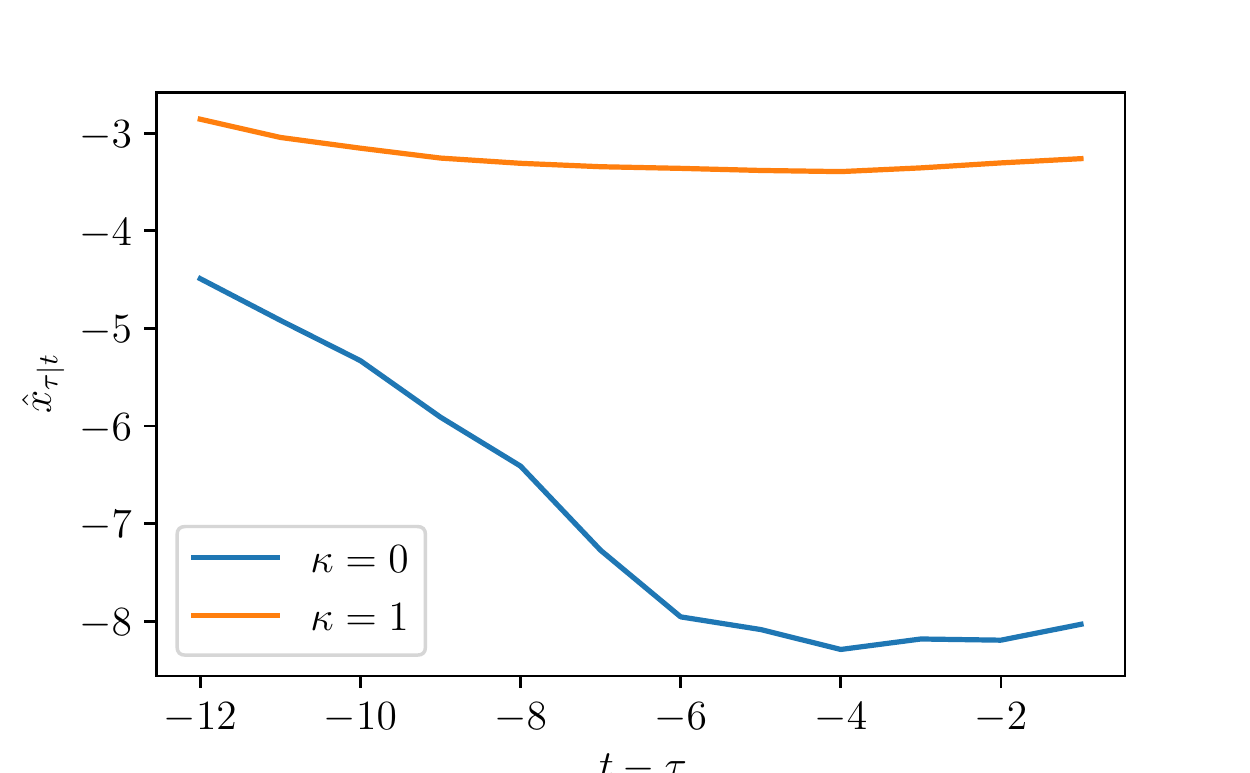}
\caption{Simulated dataset. Forecasts of $(x_\tau)_1$ made at $t=\tau-1,\ldots,\tau-H$ for two forecasters:
one with $\kappa=0$ and one with $\kappa=1$. Notice that the forecasts are much more consistent and barely change when
$\kappa=1$.}
\label{fig:simulated_consistency}
\end{figure}

We can trade off forecaster consistency for performance.
We fit the forecaster with $\alpha=0.1$ for a number of values of
$\kappa\in[10^{-2}, 10^1]$.
In figure \ref{fig:simulated_consistency_vs_test_loss} we compare the
train and test loss versus the train and test forecaster consistency
for these values of $\kappa$.
To improve forecaster consistency, we have to sacrifice performance
(\eg, a 1000x reduction in forecaster consistency more than doubles our test loss).
Finally, in figure \ref{fig:simulated_consistency} we demonstrate the effect of
encouraging forecaster consistency.

\clearpage
\subsection{Stock index absolute returns}
In this example, we use previous absolute returns of a stock index
to forecast future absolute returns of the index.
It has been observed that stocks exhibit volatility clustering,
first observed by Mandelbrot when he wrote
``large changes tend to be followed by large changes, of either sign,
and small changes tend to be followed by small changes'' \cite{mandelbrot_variation_1963}.
In this example we use the techniques of low rank forecasting to
analyze volatility clustering.
Along the way, we find that the latent state in a rank one forecaster
very closely resembles the CBOE Volatility index (VIX),
an index that tracks the 30-day expected volatility of the US stock market.
We note that our model is very similar in spirit to a GARCH model, which
has been observed to work well for modeling market volatility \cite{ding_modeling_1996}.

We gathered the daily absolute return of the SPY ETF (exchange traded fund),
an ETF that closely tracks the S\&P 500 index,
from February 1993 to October 2020 ($T\approx 7000$).
We annualize the daily absolute return by multiplying them by $\sqrt{250}$.
We split the original dataset in half, into a training and test dataset.
We also pre-process both datasets by subtracting the mean of $\{x_t\}$
on the training dataset from both the training and test dataset.

Our goal will be to predict the next month of SPY's absolute returns ($H=20$ trading days)
from the past quarter of SPY's absolute returns ($M=60$ trading days).
The parameter $\theta$ is thus a 60 by 20 matrix, containing $1200$ entries.

\begin{figure}
\centering
\includegraphics[]{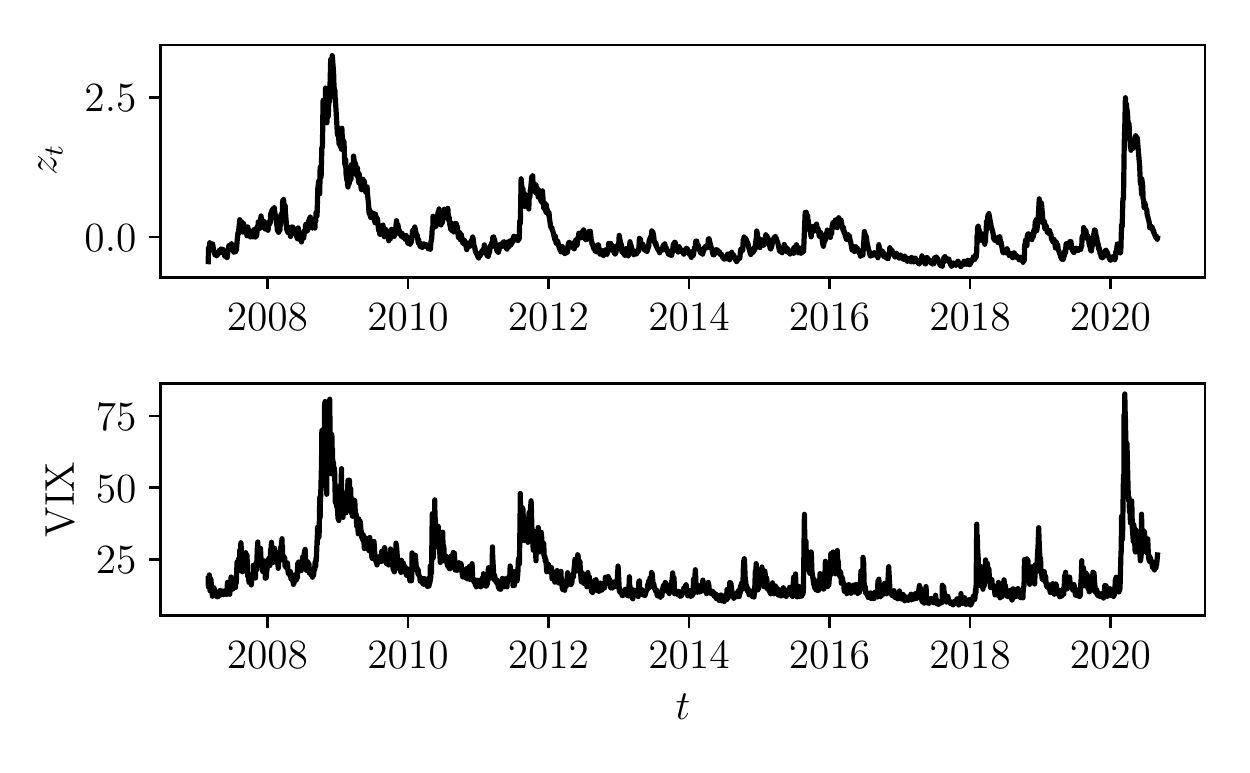}
\caption{\emph{Top.} Latent state in rank one forecaster on (unseen) test set. \emph{Bottom.}
VIX index.}
\label{fig:spy_latent_state}
\end{figure}

\begin{figure}
\centering
\includegraphics[]{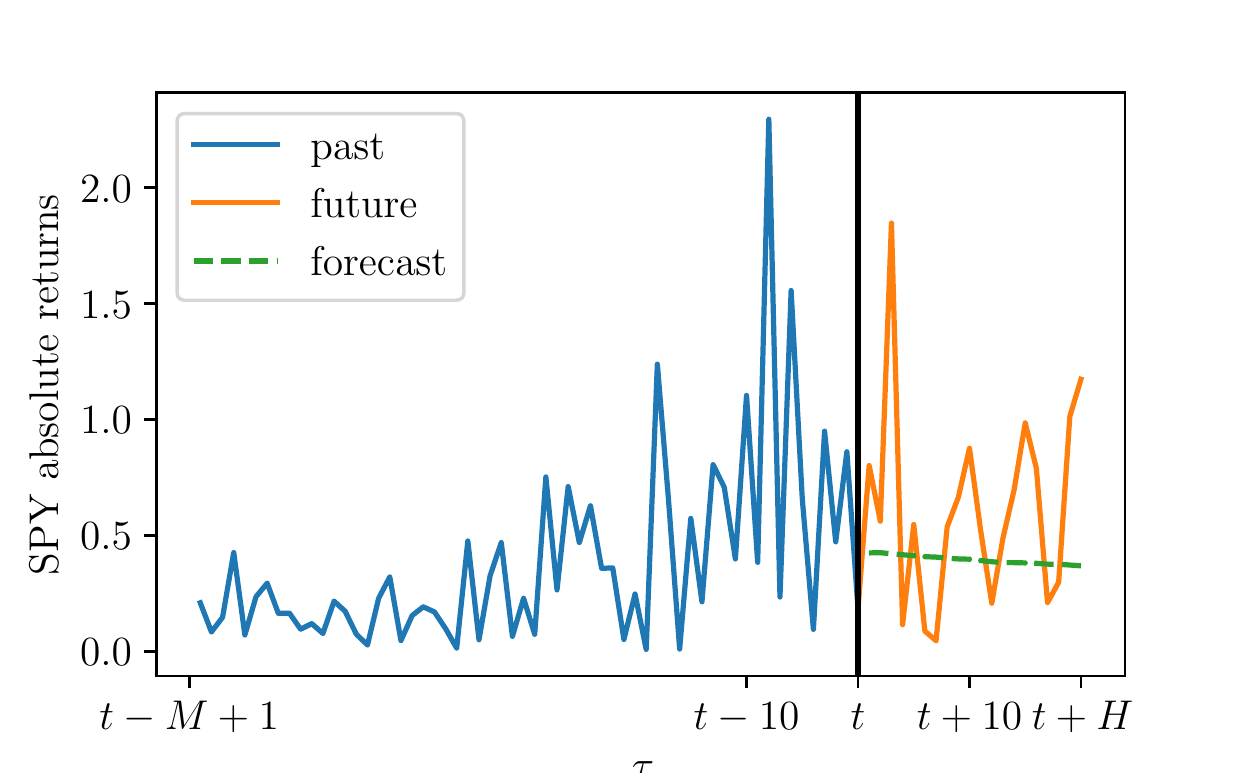}
\caption{SPY absolute return forecast.}
\label{fig:stock_forecast}
\end{figure}

\begin{figure}
\centering
\includegraphics[]{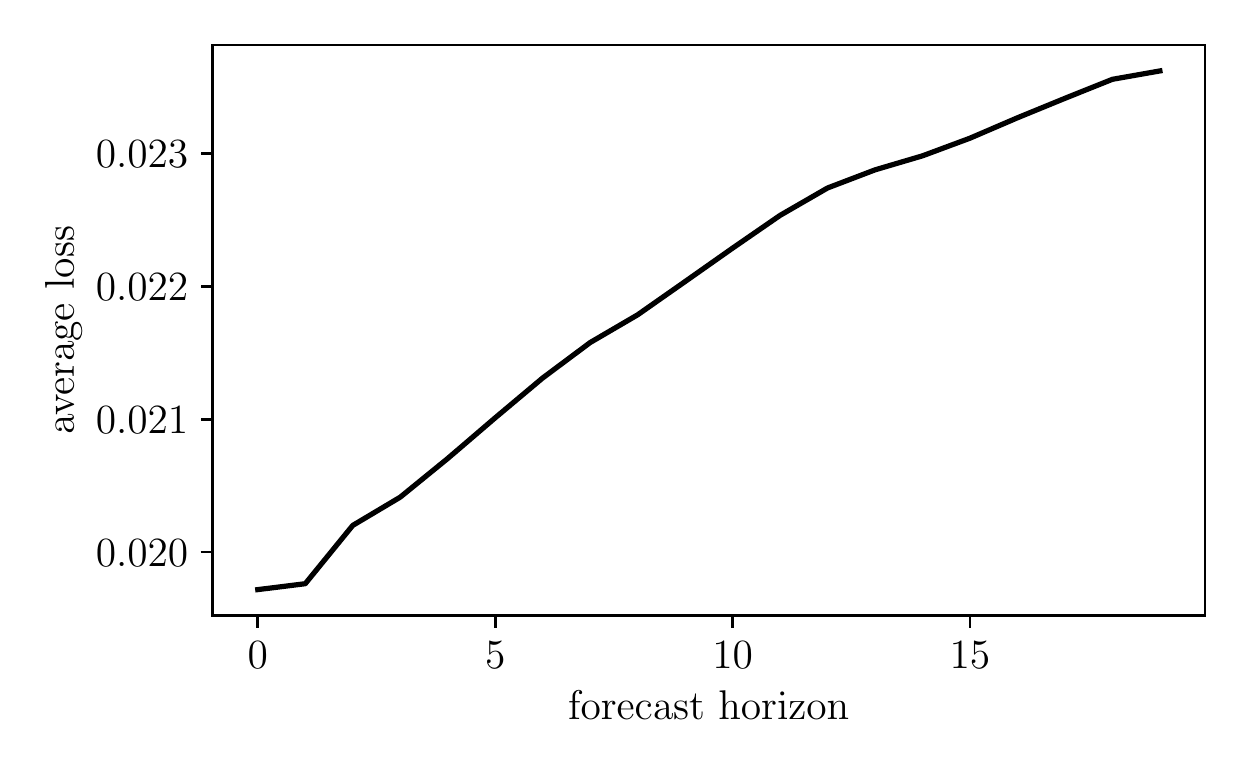}
\caption{Test loss at different forecast horizons in forecasting SPY absolute returns.}
\label{fig:stock_loss_horizon}
\end{figure}

We tried a number of values of $\alpha$, and found that $\alpha=0.05$ worked
well, and corresponds to a rank one forecaster.
Fitting each forecaster took roughly four seconds, not using warm-start.
(Warm starting the optimization from the previous value of $\alpha$ reduced this
considerably.)
The test loss of this forecaster is 0.022; for comparison,
the test loss for the mean forecaster is 0.026,
meaning our forecaster provides a 15\% percent improvement in test loss.
In figure \ref{fig:stock_forecast} we show a forecast from our model
on the test dataset.
In figure \ref{fig:stock_loss_horizon} we show the forecaster test loss
versus horizon (how many steps it is forecasting out).
As expected, the test loss increases as we forecast further out.

\begin{figure}
\centering
\includegraphics[width=0.8\textwidth]{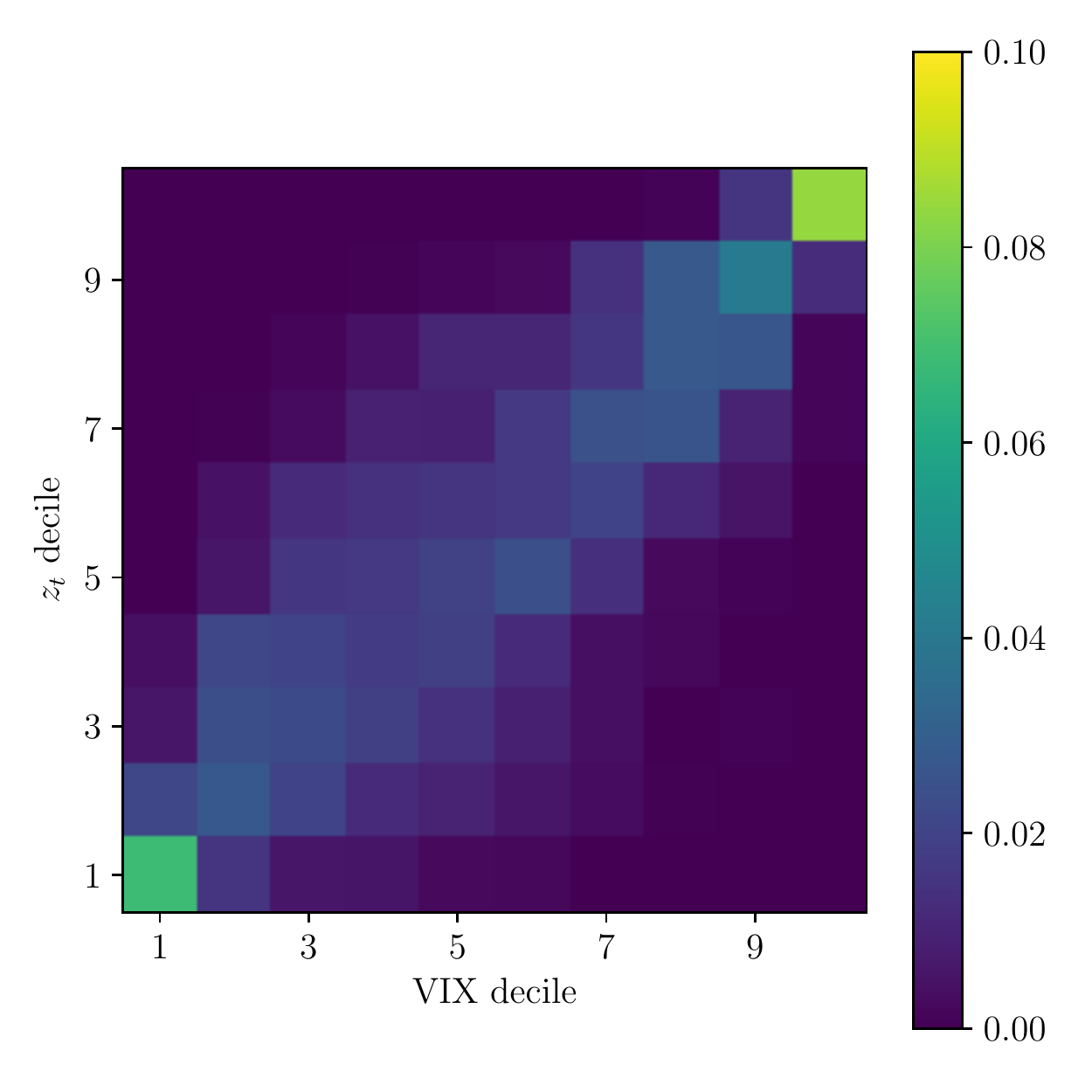}
\caption{Heatmap of $z_t$ and VIX deciles.}
\label{fig:stock_decile}
\end{figure}

In figure \ref{fig:spy_latent_state} we plot the latent state of the forecaster,
along with the VIX, on the test set.
At least visually, the forecaster's latent state looks like a (shifted and scaled) smooth version
of the VIX.
To investigate the correlation between the forecaster latent state and VIX,
we took the deciles of both series, and calculated the number of times each series were
in their respective deciles.
In figure \ref{fig:stock_decile} we show a heatmap of the deciles.
When the VIX index is in its highest decile, the forecaster latent state
is also in its highest decile over 80\% of the time.
The heatmap looks very close to diagonal, so we can conclude that the 
two time series are very correlated.

\begin{figure}
\centering
\includegraphics[]{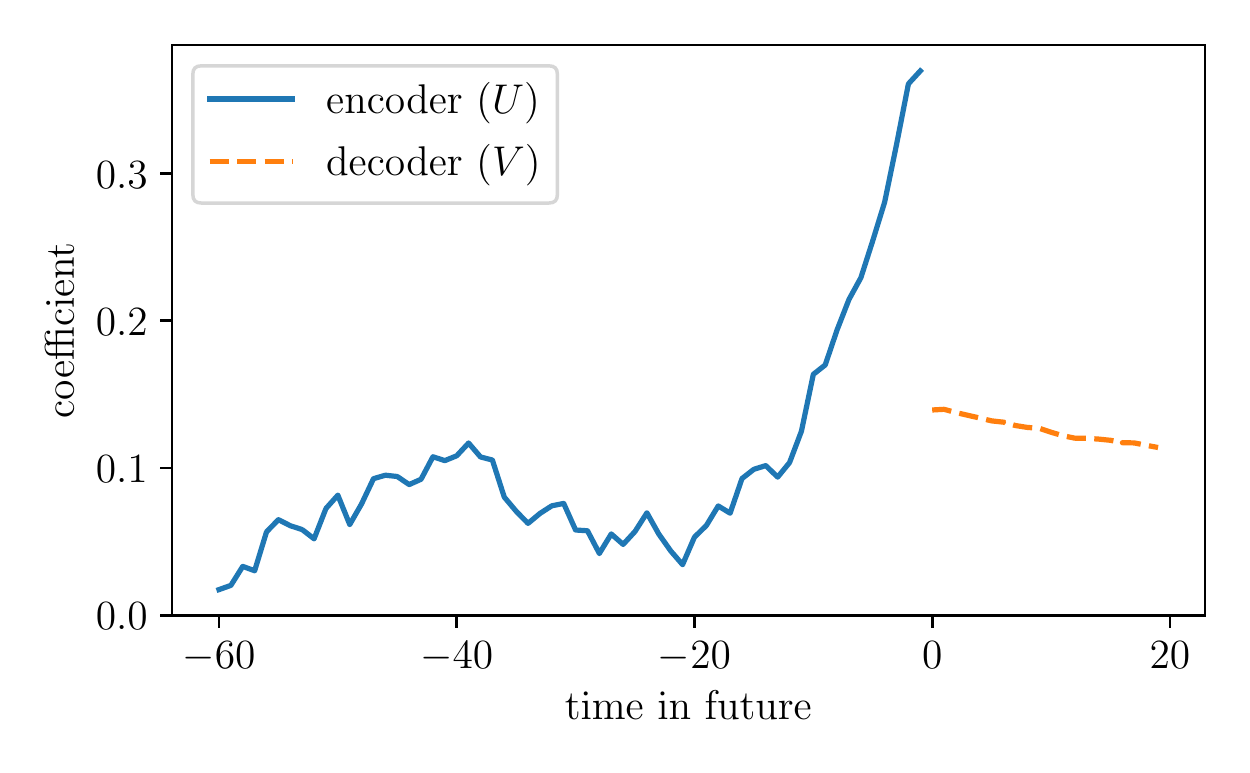}
\caption{SPY forecaster factors. We filter the last 60 days of absolute returns
using $U$ to get a single latent state, and then multiply the latent state by $V$
to get our forecast.}
\label{fig:stock_uv}
\end{figure}

Since the encoder $U$ and the decoder $V$ are both vectors
(indeed one-dimensional filters), we can plot them.
In figure \ref{fig:stock_uv} we show the parameters of the encoder and decoder.
We can see that the decoder is most sensitive to the most recent values of the series,
and also assigns high importance to the absolute returns 40 trading days ago (around $i=20$).
We also observe that the decoder $V$ is roughly decreasing in $i$; this means
that the farther out we are forecasting, the closer the forecast gets to 0, or to simply
predicting the mean value.

\clearpage
\subsection{Traffic}
We consider a dataset from the Caltrans performance measurement system (PeMS),
which is composed of hourly road occupancy rates
at $n=100$ stations located on highways in Caltrans District 4 (the San Francisco Bay Area)
from October 2019 to December 2019 \cite{choe2001freeway} ($T=2000$).
Each data entry is the average occupancy rate over the hour, between 0 and 1, which is roughly the average
fraction of the time each vehicle was present in that segment over 30 second windows (for more details
see \cite{choe2001freeway}).

\paragraph{Pre-processing.}
We carried out several pre-processing steps.
We first clip or Winsorize the raw occupancy rates $o_t$ to be in $[0.001,0.999]$,
and then perform a logit transform, \ie, 
\[
\log\left(\frac{o_t}{1-o_t}\right),
\]
where division and $\log$ are elementwise.
(This yields a more normal distribution.)

We split the dataset in half into a training and test dataset.
We subtracted the mean occupancy rate on the training dataset for each station from the
occupancy rates in both the training and test dataset.
We take $M=24$ and $H=6$, so from the last day of traffic we predict
the next quarter day of traffic.
In total, there are around 1.4 million parameters in a linear forecaster.
We use the $\ell_2$ (squared) loss.
The loss of the constant (mean) forecaster
was 1.108 on the training dataset and 0.962 on the test dataset.

\paragraph{De-trending.}
We de-trend the time series as described in \S\ref{s-auxiliary-data}
by using auxiliary time-based features.
As auxiliary features we use sine and cosine of hour
(with periods of 24, 12, 8, 6, and 24/5 hours),
hour in week (at harmonic periods of 168, 168/2, 168/3, 168/4, and 168/5 hours),
and a binary weekday/weekend feature.
We also use all pairwise products of these auxiliary features,
which for products of sines and cosines are sinusoids at the sum and difference
frequencies.
In total there are 462 auxiliary features.
This (deterministic) baseline model yields a training loss of 0.099
and a test loss of 0.203.

\begin{figure}
\centering
\includegraphics[]{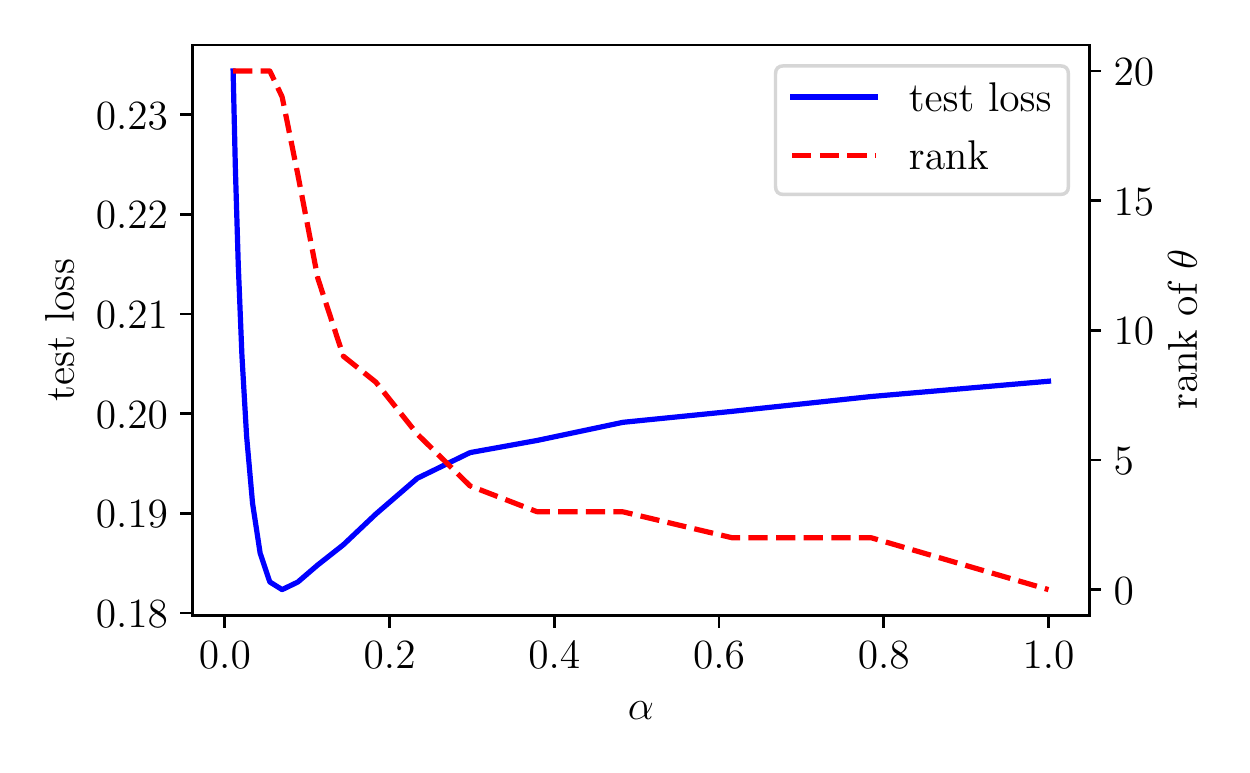}
\caption{Test loss and rank of $\theta$ versus $\alpha$ on the traffic dataset.}
\label{fig:traffic_test_loss}
\end{figure}

\begin{figure}
\centering
\includegraphics[]{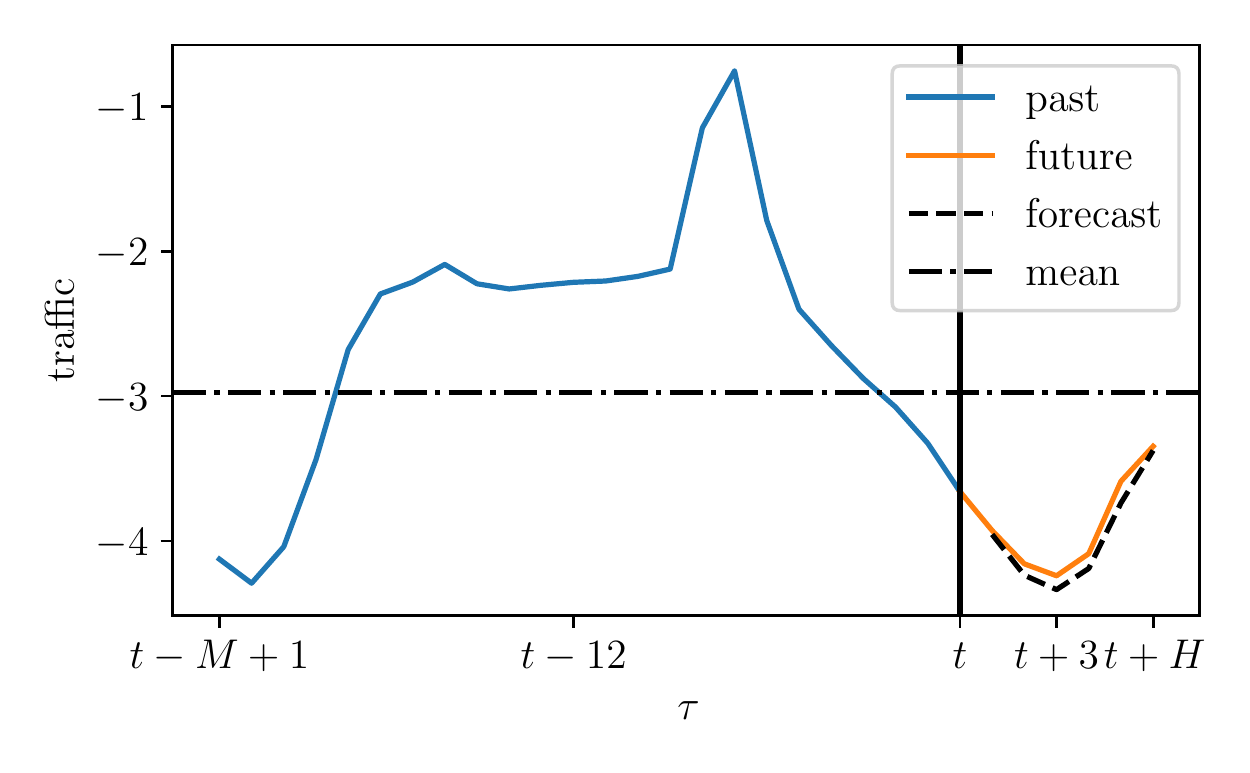}
\caption{Forecast of traffic data made at a time $t$, shown as vertical line.
The solid curve shows $(x_\tau)_1$, and the dashed
line is our forecast of the logit occupancy,
$(\hat x_{\tau\mid t})_1$, $\tau=t+1,\ldots,t+H$.}
\label{fig:traffic_forecast}
\end{figure}

\paragraph{Forecasting.}
Next, we construct a low rank forecaster for the residual series.
Figure \ref{fig:traffic_test_loss} shows the rank and test loss 
versus the regularization parameter $\alpha$.
The choice $\alpha = 0.07$ yields a rank 14 forecaster with a test loss of 0.184,
and improvement of around 10\% over the baseline model.
Fitting the forecaster takes around 11 seconds.
An example forecast is shown in figure \ref{fig:traffic_forecast}.

\clearpage
\section{Related work}
\label{s-related-work}

The general problem of time series forecasting has been studied for decades,
and many books have been written on the subject
\cite{chatfield_time-series_2000,lutkepohl_new_2005,de_gooijer_25_2006,box_time_2008,
brockwell_introduction_2010,granger_forecasting_2014,hamilton_time_2020}.
Many methods have been developed, and we survey a subset of them below.
We try to focus on methods that are by our definition low rank, \ie,
the forecast can be decomposed as first computing a lower dimensional vector (the latent state)
from the past, and then computing the forecast based on that low dimensional vector.
We also survey general work on general statistical modeling of time series and low rank
decompositions.

\paragraph{Classical time series models.}
Classical models of time series model them as stationary stochastic processes.
Given the random process, the optimal (in terms of mean squared loss) forecaster is
the conditional mean forecaster, \ie, the forecast of future values is the mean
of the future values given the observed past \cite[\S5]{box_time_2008}.
Some common linear models that fall under this category are
the Wiener filter \cite{wiener1950extrapolation},
exponential smoothing \cite{brown_exponential_1957,holt_forecasting_1957},
moving average \cite[\S3.3]{box_time_2008}, autoregressive \cite[\S3.2]{box_time_2008},
ARMA \cite[\S3.4]{box_time_2008}, and ARIMA \cite[\S4]{box_time_2008}.
Most of these models are Markov processes, meaning the future is independent of
the past, given a particular state or summarization of the past \cite{markov1954theory}.
Most of these models are also special cases of linear state space models \cite{kalman_new_1960},
and exact forecasting can be done by first Kalman filtering, and then
iterating the state space model without disturbances \cite[\S5.5]{box_time_2008}.
There exist many methods to fit state space models to data \cite[\S7]{box_time_2008}, including
N4SID \cite{van_overschee_n4sid_1994}, EM \cite{shumway_approach_1982,holmes2012derivation},
and least squares auto-tuning \cite{barratt_fitting_2020}.
State space models are very closely related to dynamic factor models
\cite{bernanke2005measuring,forni2005generalized,stock2006forecasting,bai2007determining,stock2011dynamic,
choi_efficient_2012,pena2019forecasting}, which have found applications in
economics \cite{geweke1977dynamic,figlewski1983optimal,figlewski1983optimal1,quah1993dynamic,stock1999forecasting,
stock2002forecasting,bernanke2003monetary,crone2005consistent,boivin2006more}.
Our paper is focused on the problem of forecasting, and does not explicitly construct
a stochastic model of the time series.

\paragraph{Minimum order system identification.}
A closely related problem in system identification is
finding the minimal order representation of a linear system \cite{gilbert1963controllability,kalman1963mathematical}.
One approach is to construct a Hankel matrix of impulse responses, which was
first proposed by Ho and Kalman \cite{ho1966effective}, and later expanded upon by Tether \cite{tether1970construction}, Risannen
\cite{rissanen1971recursive}, Woodside \cite{woodside1971estimation}, Silverman \cite{silverman1971realization},
Akaike \cite{akaike1974stochastic}, Chow \cite{chow1972estimating,chow1972estimation}, and Aoki \cite{aoki1983estimation,aoki1983prediction}.
These early papers spawned the field of subspace identification \cite{katayama2006subspace,van2012subspace},
which has resulted in techniques like N4SID \cite{moonen1989and,van1992two,van1993subspace,van_overschee_n4sid_1994}, which take the SVD of a particular block Hankel matrix,
and related techniques like
MOESP \cite{verhaegen1991novel,verhaegen1992subspace,verhaegen1993subspace} and
CVA \cite{larimore1983system,larimore1990canonical} (see, \eg, \cite[\S10.6]{ljung1999system} for a summary).
We also point the reader to the paper by Jansson \cite{jansson1996linear}, which poses
subspace identification methods as a regression, where the forecasting matrix is low rank.

\paragraph{Low rank matrix approximation.}
The techniques in this paper are closely related
to the formation of low rank approximations to matrices,
which dates all the way back to Eckart's seminal work on
principal component analysis (PCA) \cite{eckart1936approximation}.
The basis of PCA is that the truncated SVD of a matrix
is the best low rank approximation of the matrix (in terms of Frobenius norm), and
has been expanded heavily and generalized to different data types,
objective functions, and regularization functions \cite{chu2003structured,udell_generalized_2016}.
A standard convex, continuous approximation of the rank function
is the nuclear norm function \cite{fazel_rank_2001},
and convex optimization problems involving nuclear norms can be
expressed as semidefinite programs \cite{vandenberghe1996semidefinite},
and efficient solution methods exist (see, \eg, \cite{liu_interior-point_2009,cai2010singular}).
The nuclear norm also exhibits a number of nice theoretical properties,
\eg, it can be used to recover the true minimum rank solution for certain problems
\cite{recht_guaranteed_2010}, and near-optimal solutions in others \cite{candes2010power}.
It has also proved to be a useful heuristic for system identification problems
\cite{fazel_rank_2001,mohan2010reweighted,hansson_subspace_2012,fazel2013hankel}.
For more discussion on the nuclear norm, and its applications, see, \eg,
\cite{pong_trace_2010,negahban_estimation_2011,ma_fixed_2011,markovsky2012low,markovsky2012effective,shamir2014matrix}
and the references therein.

\paragraph{Reduced rank prediction.}
A closely related problem is reduced rank prediction. Reduced rank regression can be traced 
back to the work \cite{anderson1951estimating}, where a likelihood-ratio test is obtained for the hypothesis that the rank of the 
regression coefficient matrix is a given number. Later, the work \cite{izenman1975reduced} provided an explicit form of the 
estimate of the regression coefficient matrix with a given rank, and discussed the asymptotics of the estimated 
regression coefficient matrix. The reduced rank regression problem was approximately solved as a nuclear norm penalized least 
squares problem in \cite{yuan_dimension_2007}, and then further generalized into an 
adaptive nuclear norm penalized reduced rank regression problem in \cite{chen2013reduced}.
For more details on reduced rank regression, please refer to \cite{velu2013multivariate}.
The optimization problem in the 
form of a loss function plus a nuclear norm regularization term has applications in
in multi-task learning; see, \eg, \cite{caruana_multitask_1998,amit_uncovering_2007,argyriou2008convex,pong_trace_2010}.

\paragraph{Reduced rank time series.}
Similar to reduced rank regression, reduced rank time series modeling is also closely connected with the low rank 
forecasting problem. Early work such as \cite{ahn_nested_1988,velu_reduced_1986,wang_forecasting_2004} fit reduced 
rank coefficient models to vector time series to provide a concise representation of vector time series models.
Many of these models work by 
extracting a lower dimensional vector time series from the past first, and then perform one-step ahead forecast 
(which corresponds to $H=1$) based on the extracted low dimensional vector time series. This is very similar to the (low rank) two step 
forecasting technique discussed in this paper. The reduced rank modeling problem has been generalized into the structured 
VAR modeling problem \cite{basu_low_2019,alquier_high-dimensional_2020,melnyk2016estimating}. In these papers, regularization terms 
are added to encourage certain structures. For example, a nuclear norm term is often used to encourage the transition matrix 
to be low rank, and $\ell_1$ norm is used to encourage the transition matrix to be sparse.

\paragraph{Extracting a low-dimensional predictable time series.}  
Another related problem is extract a low-dimensional (self-)predictable
time series from a high-dimensional vector time series.
The paper \cite{box1977canonical} is an early one with
the explicit goal of predictability.  After this,
many other methods have been developed in different research areas on
extracting predictable time series, including economics
\cite{box1977canonical,pena_identifying_1987,lam_estimation_2011,lam2012factor,
dong_extracting_2020}, machine learning
\cite{stone2001blind,richthofer2015predictable,goerg2013forecastable}, process
system engineering
\cite{li2014new,zhou2016autoregressive,dong_dynamic_2018,dong2018novel}, signal
processing, and atmospheric research
\cite{delsole2001optimally,delsole2009average1,delsole2009average2}.  
Another closely related line of work is on predictive state representations
\cite{littman_predictive_2002,rosencrantz2004learning,boots2011closing}. 

\paragraph{Nonlinear forecasters.}
There also exist many
nonlinear forecasting methods. These methods are often based on neural networks, \eg,
recurrent neural networks \cite{connor1994recurrent,che2018recurrent,maggiolo2019autoregressive,siami2018forecasting,
gers2002applying,chen2015lstm}, convolutional neural networks \cite{livieris2020cnn,afrasiabi2019dtw,li2020real}, and autoencoders 
\cite{bao2017deep,gensler2016deep,wei2019autoencoder}. Other nonlinear forecasting methods include regime switching models \cite{sola1994testing,nyberg2018forecasting} and NARMAX models \cite{johansen1993constructing,chen1989representations}.
Recently, there have also been methods proposed that perform vector time series 
forecasting through matrix completion, where the vector time series are assumed share some common structures \cite{gillard2018structured,chen2020low}.

\section*{Acknowledgements}

The authors gratefully acknowledge conversations and discussions about 
some of the material in this paper with Enzo Busseti, Misha van Beek, and Linxi Chen.

\clearpage
\bibliography{citations}

\clearpage
\appendix
\section{Gradients}
\label{s-gradients}

\paragraph{Loss plus nuclear norm.}
Let 
\[
\mathcal L(U,V) = \frac{1}{N} \ones^T \ell(PUV-F) + \frac{\lambda}{2}(\|U\|_F^2 + \|V\|_F^2)
\]
denote the loss plus nuclear norm part of the objective in problem \eqref{eq:alternating}.
Let
\[
g_i = \nabla_{\hat f_i} \ell(\hat f_i) \in \reals^{Hn}, \quad i=1,\ldots,N
\]
denote the gradient of the loss with respect to the forecast of the $i$th data point.
We have that
\[
\nabla_U \mathcal L(U,V) = \frac{1}{N} \sum_{i=1}^N p_ig_i^TV^T + \lambda U,\qquad
\nabla_V \mathcal L(U,V) = \frac{1}{N} \sum_{i=1}^N U^Tp_i g_i^T + \lambda V.
\]

\paragraph{Forecaster consistency.}
To compute the gradient of the forecaster consistency loss
with respect to $U$ and $V$,
we first compute the gradient with respect to $\theta$, denoted
\[
G = \nabla \mathcal I_\theta(\theta) = P^T(\nabla_{\hat F} \dist(\hat F)^2).
\]
The matrix $\nabla_{\hat F} \dist(\hat F)^2$ is the gradient of the forecaster consistency
with respect to the forecasts $\hat F$, and its (block) elements are given by
\[
g_{\tau \mid t} = \nabla_{x_{\tau \mid t}} \dist(\hat F)^2 = 2(\hat x_{\tau \mid t} - \bar x_\tau) - \sum_{t_1=\max\{\tau-H,M+1\}}^{\min\{\tau-1,T-H\}} \frac{2}{f(\tau)}(\hat x_{\tau \mid t_1} - \bar x_\tau),
\]
where
\[
f(\tau) = \min\{\tau,T-H+1\}-\max\{\tau-H,M+1\}, \quad \tau=M+1,\ldots,T.
\]
Then, using $G$, the gradients of forecaster consistency with respect to $U$ and $V$ are
\[
\nabla_U \mathcal I (UV) = GV^T, \qquad \nabla_V \mathcal I(UV) = U^TG.
\]

\end{document}

%% file: defs.tex
\newcommand{\ones}{\mathbf 1}
\newcommand{\reals}{{\mbox{\bf R}}}

\newcommand{\rank}{\mathop{\bf rank}}

\newcommand{\Expect}{\mathop{\bf E{}}}

\newcommand{\dist}{\mathop{\bf dist{}}}

\newcommand{\eg}{{\it e.g.}}
\newcommand{\ie}{{\it i.e.}}

\newcommand{\BEAS}{\begin{eqnarray*}}
\newcommand{\EEAS}{\end{eqnarray*}}
\newcommand{\BEA}{\begin{eqnarray}}
\newcommand{\EEA}{\end{eqnarray}}
\newcommand{\BEQ}{\begin{equation}}
\newcommand{\EEQ}{\end{equation}}
\newcommand{\BIT}{\begin{itemize}}
\newcommand{\EIT}{\end{itemize}}

%% file: low_rank_forecasting.bbl
\begin{thebibliography}{100}

\bibitem{kalman_new_1960}
Rudolf Kalman.
\newblock A new approach to linear filtering and prediction problems.
\newblock {\em Journal Basic Engineering}, 82(1):35--45, 1960.

\bibitem{boyd_convex_2004}
Stephen Boyd and Lieven Vandenberghe.
\newblock {\em Convex optimization}.
\newblock Cambridge University Press, 2004.

\bibitem{hilary2013analyst}
Gilles Hilary and Charles Hsu.
\newblock Analyst forecast consistency.
\newblock {\em Journal of Finance}, 68(1):271--297, 2013.

\bibitem{gubner2006probability}
John Gubner.
\newblock {\em Probability and random processes for electrical and computer
  engineers}.
\newblock Cambridge University Press, 2006.

\bibitem{box_time_2008}
George Box, Gwilym Jenkins, and Gregory Reinsel.
\newblock {\em Time series analysis: forecasting and control}.
\newblock Wiley, 2008.

\bibitem{hastie2009elements}
Trevor Hastie, Robert Tibshirani, and Jerome Friedman.
\newblock {\em The elements of statistical learning: data mining, inference,
  and prediction}.
\newblock Springer Science \& Business Media, 2009.

\bibitem{pourahmadi2013high}
Mohsen Pourahmadi.
\newblock {\em High-dimensional covariance estimation: with high-dimensional
  data}.
\newblock John Wiley \& Sons, 2013.

\bibitem{burg1968new}
John Burg.
\newblock A new analysis technique for time series data.
\newblock {\em NATO Advanced Study Institute on Signal Processing}, 1968.

\bibitem{gilbert1963controllability}
Elmer Gilbert.
\newblock Controllability and observability in multivariable control systems.
\newblock {\em Journal of the Society for Industrial and Applied Mathematics,
  Series A: Control}, 1(2):128--151, 1963.

\bibitem{kalman1963mathematical}
Rudolf Kalman.
\newblock Mathematical description of linear dynamical systems.
\newblock {\em Journal of the Society for Industrial and Applied Mathematics,
  Series A: Control}, 1(2):152--192, 1963.

\bibitem{aoki2013state}
Masanao Aoki.
\newblock {\em State space modeling of time series}.
\newblock Springer Science \& Business Media, 2013.

\bibitem{van_overschee_n4sid_1994}
Peter Van~Overschee and Bart De~Moor.
\newblock {N4SID}: {Subspace} algorithms for the identification of combined
  deterministic-stochastic systems.
\newblock {\em Automatica}, 30(1):75--93, 1994.

\bibitem{shumway_approach_1982}
Robert Shumway and David Stoffer.
\newblock An approach to time series smoothing and forecasting using the {EM}
  algorithm.
\newblock {\em Journal Time Series Analysis}, 3(4):253--264, 1982.

\bibitem{barratt_fitting_2020}
Shane Barratt and Stephen Boyd.
\newblock Fitting a {Kalman} smoother to data.
\newblock In {\em American {Control} {Conference} ({ACC})}, pages 1526--1531.
  IEEE, 2020.

\bibitem{boyd_introduction_2018}
Stephen Boyd and Lieven Vandenberghe.
\newblock {\em Introduction to applied linear algebra: vectors, matrices, and
  least squares}.
\newblock Cambridge University Press, June 2018.

\bibitem{diamond_cvxpy_2016}
Steven Diamond and Stephen Boyd.
\newblock {CVXPY}: A {P}ython-embedded modeling language for convex
  optimization.
\newblock {\em Journal of Machine Learning Research}, 17(83):1--5, 2016.

\bibitem{agrawal_rewriting_2018}
Akshay Agrawal, Robin Verschueren, Steven Diamond, and Stephen Boyd.
\newblock A rewriting system for convex optimization problems.
\newblock {\em Journal of Control and Decision}, 5(1):42--60, 2018.

\bibitem{ji_accelerated_2009}
Shuiwang Ji and Jieping Ye.
\newblock An accelerated gradient method for trace norm minimization.
\newblock In {\em Proc. {Intl.} {Conf.} {Machine} {Learning}}, pages 457--464,
  2009.

\bibitem{liu_nuclear_2014}
Yuanyuan Liu, Hong Cheng, Fanhua Shang, and James Cheng.
\newblock Nuclear norm regularized least squares optimization on {Grassmannian}
  manifolds.
\newblock In {\em Proc. {Uncertainty} in {Artificial} {Intelligence}}, pages
  515--524. AUAI Press, 2014.

\bibitem{toh_accelerated_nodate}
Kim-Chuan Toh and Sangwoon Yun.
\newblock An accelerated proximal gradient algorithm for nuclear norm
  regularized linear least squares problems.
\newblock {\em Pacific Journal of Optimization}, 6(615-640):15, 2010.

\bibitem{keshavan2012efficient}
Raghunandan Keshavan.
\newblock {\em Efficient algorithms for collaborative filtering}.
\newblock PhD thesis, Stanford University, 2012.

\bibitem{chen2012matrix}
Caihua Chen, Bingsheng He, and Xiaoming Yuan.
\newblock Matrix completion via an alternating direction method.
\newblock {\em IMA Journal of Numerical Analysis}, 32(1):227--245, 2012.

\bibitem{jain2013low}
Prateek Jain, Praneeth Netrapalli, and Sujay Sanghavi.
\newblock Low-rank matrix completion using alternating minimization.
\newblock In {\em Proc. ACM Symposium on Theory of Computing}, pages 665--674,
  2013.

\bibitem{hardt2013provable}
Moritz Hardt.
\newblock On the provable convergence of alternating minimization for matrix
  completion.
\newblock {\em arXiv preprint arXiv:1312.0925}, 2013.

\bibitem{netrapalli2014non}
Praneeth Netrapalli, UN~Niranjan, Sujay Sanghavi, Animashree Anandkumar, and
  Prateek Jain.
\newblock Non-convex robust pca.
\newblock In {\em Advances in Neural Information Processing Systems}, pages
  1107--1115, 2014.

\bibitem{agarwal2016learning}
Alekh Agarwal, Animashree Anandkumar, Prateek Jain, and Praneeth Netrapalli.
\newblock Learning sparsely used overcomplete dictionaries via alternating
  minimization.
\newblock {\em SIAM Journal on Optimization}, 26(4):2775--2799, 2016.

\bibitem{udell_generalized_2016}
Madeleine Udell, Corinne Horn, Reza Zadeh, and Stephen Boyd.
\newblock Generalized low rank models.
\newblock {\em Foundations and Trends® in Machine Learning}, 9(1):1--118,
  2016.

\bibitem{liu_limited_1989}
Dong Liu and Jorge Nocedal.
\newblock On the limited memory {BFGS} method for large scale optimization.
\newblock {\em Mathematical Programming}, 45(1-3):503--528, 1989.

\bibitem{goodfellow_deep_2016}
Ian Goodfellow, Yoshua Bengio, and Aaron Courville.
\newblock {\em Deep learning}.
\newblock MIT Press, 2016.

\bibitem{velasquez-bermudez_dynamic_2019}
Nicholas Moehle, Enzo Busseti, Stephen Boyd, and Matt Wytock.
\newblock Dynamic energy management.
\newblock In {\em Large {Scale} {Optimization} in {Supply} {Chains} and {Smart}
  {Manufacturing}}, volume 149, pages 69--126. Springer International
  Publishing, 2019.

\bibitem{paszke_pytorch_2019}
Adam Paszke, Sam Gross, Francisco Massa, Adam Lerer, James Bradbury, Gregory
  Chanan, Trevor Killeen, Zeming Lin, Natalia Gimelshein, Luca Antiga, and
  {others}.
\newblock {PyTorch}: {An} imperative style, high-performance deep learning
  library.
\newblock In {\em Advances in {Neural} {Information} {Processing} {Systems}},
  pages 8024--8035, 2019.

\bibitem{mandelbrot_variation_1963}
Benoit Mandelbrot.
\newblock The variation of certain speculative prices.
\newblock {\em The Journal of Business}, 36(4):394--419, 1963.

\bibitem{ding_modeling_1996}
Zhuanxin Ding and Clive W.~J. Granger.
\newblock Modeling volatility persistence of speculative returns: {A} new
  approach.
\newblock {\em Journal of Econometrics}, 73(1):185--215, July 1996.

\bibitem{choe2001freeway}
Tom Choe, Alexander Skabardonis, and Pravin Varaiya.
\newblock Freeway performance measurement system ({PeMS}): An operational
  analysis tool.
\newblock {\em 81st Annual Meeting, Transportation Research Board}, 2001.

\bibitem{chatfield_time-series_2000}
Chris Chatfield.
\newblock {\em Time-series forecasting}.
\newblock CRC press, 2000.

\bibitem{lutkepohl_new_2005}
Helmut L{\"u}tkepohl.
\newblock {\em New introduction to multiple time series analysis}.
\newblock Springer Science \& Business Media, 2005.

\bibitem{de_gooijer_25_2006}
Jan De~Gooijer and Rob Hyndman.
\newblock 25 years of time series forecasting.
\newblock {\em International Journal of Forecasting}, 22(3):443--473, 2006.

\bibitem{brockwell_introduction_2010}
Peter Brockwell, Richard Davis, and Matthew Calder.
\newblock {\em Introduction to time series and forecasting}, volume~2.
\newblock Springer, 2002.

\bibitem{granger_forecasting_2014}
John Granger and Paul Newbold.
\newblock {\em Forecasting economic time series}.
\newblock Academic Press, 2014.

\bibitem{hamilton_time_2020}
James Hamilton.
\newblock {\em Time series analysis}.
\newblock Princeton University Press, 2020.

\bibitem{wiener1950extrapolation}
Norbert Wiener.
\newblock {\em Extrapolation, interpolation, and smoothing of stationary time
  series: with engineering applications}.
\newblock MIT Press, 1950.

\bibitem{brown_exponential_1957}
Robert Brown.
\newblock Exponential smoothing for predicting demand.
\newblock In {\em Operations Research}, volume~5, pages 145--145, 1957.

\bibitem{holt_forecasting_1957}
Charles Holt.
\newblock Forecasting trends and seasonal by exponentially weighted moving
  averages.
\newblock {\em ONR Memorandum}, 52, 1957.

\bibitem{markov1954theory}
Andrei Markov.
\newblock The theory of algorithms.
\newblock {\em Trudy Matematicheskogo Instituta Imeni VA Steklova}, 42:3--375,
  1954.

\bibitem{holmes2012derivation}
Elizabeth Holmes.
\newblock Derivation of the {EM} algorithm for constrained and unconstrained
  {MARSS} models.
\newblock 2012.

\bibitem{bernanke2005measuring}
Ben Bernanke, Jean Boivin, and Piotr Eliasz.
\newblock Measuring the effects of monetary policy: a factor-augmented vector
  autoregressive ({FAVAR}) approach.
\newblock {\em The Quarterly Journal of Economics}, 120(1):387--422, 2005.

\bibitem{forni2005generalized}
Mario Forni, Marc Hallin, Marco Lippi, and Lucrezia Reichlin.
\newblock The generalized dynamic factor model: one-sided estimation and
  forecasting.
\newblock {\em Journal of the American Statistical Association},
  100(471):830--840, 2005.

\bibitem{stock2006forecasting}
James Stock and Mark Watson.
\newblock Forecasting with many predictors.
\newblock {\em Handbook of Economic Forecasting}, 1:515--554, 2006.

\bibitem{bai2007determining}
Jushan Bai and Serena Ng.
\newblock Determining the number of primitive shocks in factor models.
\newblock {\em Journal of Business \& Economic Statistics}, 25(1):52--60, 2007.

\bibitem{stock2011dynamic}
James Stock and Mark Watson.
\newblock Dynamic factor models.
\newblock {\em Oxford Handbooks Online}, 2011.

\bibitem{choi_efficient_2012}
In~Choi.
\newblock Efficient estimation of factor models.
\newblock {\em Econometric Theory}, pages 274--308, 2012.

\bibitem{pena2019forecasting}
Daniel Pe{\~n}a, Ezequiel Smucler, and Victor Yohai.
\newblock Forecasting multiple time series with one-sided dynamic principal
  components.
\newblock {\em Journal of the American Statistical Association},
  114(528):1683--1694, 2019.

\bibitem{geweke1977dynamic}
John Geweke.
\newblock The dynamic factor analysis of economic time series.
\newblock {\em Latent variables in socio-economic models}, 1977.

\bibitem{figlewski1983optimal}
Stephen Figlewski.
\newblock Optimal price forecasting using survey data.
\newblock {\em The Review of Economics and Statistics}, pages 13--21, 1983.

\bibitem{figlewski1983optimal1}
Stephen Figlewski and Thomas Urich.
\newblock Optimal aggregation of money supply forecasts: Accuracy,
  profitability and market efficiency.
\newblock {\em The Journal of Finance}, 38(3):695--710, 1983.

\bibitem{quah1993dynamic}
Danny Quah and Thomas Sargent.
\newblock A dynamic index model for large cross sections.
\newblock In {\em Business Cycles, Indicators, and Forecasting}, pages
  285--310. University of Chicago Press, 1993.

\bibitem{stock1999forecasting}
James Stock and Mark Watson.
\newblock Forecasting inflation.
\newblock {\em Journal of Monetary Economics}, 44(2):293--335, 1999.

\bibitem{stock2002forecasting}
James Stock and Mark Watson.
\newblock Forecasting using principal components from a large number of
  predictors.
\newblock {\em Journal of the American Statistical Association},
  97(460):1167--1179, 2002.

\bibitem{bernanke2003monetary}
Ben Bernanke and Jean Boivin.
\newblock Monetary policy in a data-rich environment.
\newblock {\em Journal of Monetary Economics}, 50(3):525--546, 2003.

\bibitem{crone2005consistent}
Theodore Crone and Alan Clayton-Matthews.
\newblock Consistent economic indexes for the 50 states.
\newblock {\em Review of Economics and Statistics}, 87(4):593--603, 2005.

\bibitem{boivin2006more}
Jean Boivin and Serena Ng.
\newblock Are more data always better for factor analysis?
\newblock {\em Journal of Econometrics}, 132(1):169--194, 2006.

\bibitem{ho1966effective}
Bin-Lun Ho and Rudolf Kalman.
\newblock Effective construction of linear state-variable models from
  input/output functions.
\newblock {\em at-Automatisierungstechnik}, 14(1-12):545--548, 1966.

\bibitem{tether1970construction}
Anthony Tether.
\newblock Construction of minimal linear state-variable models from finite
  input-output data.
\newblock {\em IEEE Transactions on Automatic Control}, 15(4):427--436, 1970.

\bibitem{rissanen1971recursive}
Jorma Rissanen.
\newblock Recursive identification of linear systems.
\newblock {\em SIAM Journal on Control}, 9(3):420--430, 1971.

\bibitem{woodside1971estimation}
CM~Woodside.
\newblock Estimation of the order of linear systems.
\newblock {\em Automatica}, 7(6):727--733, 1971.

\bibitem{silverman1971realization}
Leonard Silverman.
\newblock Realization of linear dynamical systems.
\newblock {\em IEEE Transactions on Automatic Control}, 16(6):554--567, 1971.

\bibitem{akaike1974stochastic}
Hirotugu Akaike.
\newblock Stochastic theory of minimal realization.
\newblock {\em IEEE Transactions on Automatic Control}, 19(6):667--674, 1974.

\bibitem{chow1972estimating}
Joseph Chow.
\newblock On estimating the orders of an autoregressive moving-average process
  with uncertain observations.
\newblock {\em IEEE Transactions on Automatic Control}, 17(5):707--709, 1972.

\bibitem{chow1972estimation}
Joseph Chow.
\newblock On the estimation of the order of a moving-average process.
\newblock {\em IEEE Transactions on Automatic Control}, 17(3):386--387, 1972.

\bibitem{aoki1983estimation}
Masanao Aoki.
\newblock Estimation of system matrices: Initial phase.
\newblock In {\em Notes on Economic Time Series Analysis: System Theoretic
  Perspectives}, pages 60--89. Springer, 1983.

\bibitem{aoki1983prediction}
Masanao Aoki.
\newblock Prediction of time series.
\newblock In {\em Notes on Economic Time Series Analysis: System Theoretic
  Perspectives}, pages 38--47. Springer, 1983.

\bibitem{katayama2006subspace}
Tohru Katayama.
\newblock {\em Subspace methods for system identification}.
\newblock Springer Science \& Business Media, 2006.

\bibitem{van2012subspace}
Peter Van~Overschee and Bart De~Moor.
\newblock {\em Subspace identification for linear systems:
  Theory-Implementation-Applications}.
\newblock Springer Science \& Business Media, 2012.

\bibitem{moonen1989and}
Marc Moonen, Bart De~Moor, Lieven Vandenberghe, and Joos Vandewalle.
\newblock On-and off-line identification of linear state-space models.
\newblock {\em International Journal of Control}, 49(1):219--232, 1989.

\bibitem{van1992two}
Peter Van~Overschee and Bart De~Moor.
\newblock Two subspace algorithms for the identification of combined
  deterministic-stochastic systems.
\newblock In {\em Proc. Conference on Decision and Control}, pages 511--516.
  IEEE, 1992.

\bibitem{van1993subspace}
Peter Van~Overschee and Bart De~Moor.
\newblock Subspace algorithms for the stochastic identification problem.
\newblock {\em Automatica}, 29(3):649--660, 1993.

\bibitem{verhaegen1991novel}
Michel Verhaegen.
\newblock A novel non-iterative {MIMO} state space model identification
  technique.
\newblock {\em IFAC Proceedings Volumes}, 24(3):749--754, 1991.

\bibitem{verhaegen1992subspace}
Michel Verhaegen and Patrick Dewilde.
\newblock Subspace model identification part 2. {A}nalysis of the elementary
  output-error state-space model identification algorithm.
\newblock {\em International Journal of Control}, 56(5):1211--1241, 1992.

\bibitem{verhaegen1993subspace}
Michel Verhaegen.
\newblock Subspace model identification part 3. {A}nalysis of the ordinary
  output-error state-space model identification algorithm.
\newblock {\em International Journal of Control}, 58(3):555--586, 1993.

\bibitem{larimore1983system}
Wallace Larimore.
\newblock System identification, reduced-order filtering and modeling via
  canonical variate analysis.
\newblock In {\em Proc. American Control Conference}, pages 445--451. IEEE,
  1983.

\bibitem{larimore1990canonical}
Wallace Larimore.
\newblock Canonical variate analysis in identification, filtering, and adaptive
  control.
\newblock In {\em Proc. Conf. Decision and Control}, pages 596--604. IEEE,
  1990.

\bibitem{ljung1999system}
Lennart Ljung.
\newblock System identification.
\newblock {\em Wiley Encyclopedia of Electrical and Electronics Engineering},
  pages 1--19, 1999.

\bibitem{jansson1996linear}
Magnus Jansson and Bo~Wahlberg.
\newblock A linear regression approach to state-space subspace system
  identification.
\newblock {\em Signal Processing}, 52(2):103--129, 1996.

\bibitem{eckart1936approximation}
Carl Eckart and Gale Young.
\newblock The approximation of one matrix by another of lower rank.
\newblock {\em Psychometrika}, 1(3):211--218, 1936.

\bibitem{chu2003structured}
Moody Chu, Robert Funderlic, and Robert Plemmons.
\newblock Structured low rank approximation.
\newblock {\em Linear Algebra and Its Applications}, 366:157--172, 2003.

\bibitem{fazel_rank_2001}
Maryam Fazel, Haitham Hindi, and Stephen Boyd.
\newblock A rank minimization heuristic with application to minimum order
  system approximation.
\newblock In {\em Proc. American Control Conference}, volume~6, pages
  4734--4739. IEEE, 2001.

\bibitem{vandenberghe1996semidefinite}
Lieven Vandenberghe and Stephen Boyd.
\newblock Semidefinite programming.
\newblock {\em SIAM Review}, 38(1):49--95, 1996.

\bibitem{liu_interior-point_2009}
Zhang Liu and Lieven Vandenberghe.
\newblock Interior-point method for nuclear norm approximation with application
  to system identification.
\newblock {\em SIAM Journal on Matrix Analysis and Applications},
  31(3):1235--1256, 2010.

\bibitem{cai2010singular}
Jian-Feng Cai, Emmanuel Cand{\`e}s, and Zuowei Shen.
\newblock A singular value thresholding algorithm for matrix completion.
\newblock {\em SIAM Journal on Optimization}, 20(4):1956--1982, 2010.

\bibitem{recht_guaranteed_2010}
Benjamin Recht, Maryam Fazel, and Pablo Parrilo.
\newblock Guaranteed minimum-rank solutions of linear matrix equations via
  nuclear norm minimization.
\newblock {\em SIAM review}, 52(3):471--501, 2010.

\bibitem{candes2010power}
Emmanuel Cand{\`e}s and Terence Tao.
\newblock The power of convex relaxation: near-optimal matrix completion.
\newblock {\em IEEE Transactions on Information Theory}, 56(5):2053--2080,
  2010.

\bibitem{mohan2010reweighted}
Karthik Mohan and Maryam Fazel.
\newblock Reweighted nuclear norm minimization with application to system
  identification.
\newblock In {\em Proc. American Control Conference}, pages 2953--2959. IEEE,
  2010.

\bibitem{hansson_subspace_2012}
Anders Hansson, Zhang Liu, and Lieven Vandenberghe.
\newblock Subspace system identification via weighted nuclear norm
  optimization.
\newblock In {\em Proc. IEEE {Conf.} {Decision} and {Control}}, pages
  3439--3444, 2012.

\bibitem{fazel2013hankel}
Maryam Fazel, Ting~Kei Pong, Defeng Sun, and Paul Tseng.
\newblock Hankel matrix rank minimization with applications to system
  identification and realization.
\newblock {\em SIAM Journal on Matrix Analysis and Applications},
  34(3):946--977, 2013.

\bibitem{pong_trace_2010}
Ting Pong, Paul Tseng, Shuiwang Ji, and Jieping Ye.
\newblock Trace norm regularization: reformulations, algorithms, and multi-task
  learning.
\newblock {\em SIAM Journal on Optimization}, 20(6):3465--3489, 2010.

\bibitem{negahban_estimation_2011}
Sahand Negahban and Martin Wainwright.
\newblock Estimation of (near) low-rank matrices with noise and
  high-dimensional scaling.
\newblock {\em Annals of Statistics}, 39(2):1069--1097, 2011.

\bibitem{ma_fixed_2011}
Shiqian Ma, Donald Goldfarb, and Lifeng Chen.
\newblock Fixed point and {Bregman} iterative methods for matrix rank
  minimization.
\newblock {\em Mathematical Programming}, 128(1-2):321--353, June 2011.

\bibitem{markovsky2012low}
Ivan Markovsky and Konstantin Usevich.
\newblock {\em Low rank approximation}, volume 139.
\newblock Springer, 2012.

\bibitem{markovsky2012effective}
Ivan Markovsky.
\newblock How effective is the nuclear norm heuristic in solving data
  approximation problems?
\newblock {\em IFAC Proceedings Volumes}, 45(16):316--321, 2012.

\bibitem{shamir2014matrix}
Ohad Shamir and Shai Shalev-Shwartz.
\newblock Matrix completion with the trace norm: learning, bounding, and
  transducing.
\newblock {\em The Journal of Machine Learning Research}, 15(1):3401--3423,
  2014.

\bibitem{anderson1951estimating}
Theodore Anderson.
\newblock Estimating linear restrictions on regression coefficients for
  multivariate normal distributions.
\newblock {\em The Annals of Mathematical Statistics}, 22(3):327--351, 1951.

\bibitem{izenman1975reduced}
Alan Izenman.
\newblock Reduced-rank regression for the multivariate linear model.
\newblock {\em Journal of Multivariate Analysis}, 5(2):248--264, 1975.

\bibitem{yuan_dimension_2007}
Ming Yuan, Ali Ekici, Zhaosong Lu, and Renato Monteiro.
\newblock Dimension reduction and coefficient estimation in multivariate linear
  regression.
\newblock {\em Journal of the Royal Statistical Society: Series B (Statistical
  Methodology)}, 69(3):329--346, 2007.

\bibitem{chen2013reduced}
Kun Chen, Hongbo Dong, and Kung-Sik Chan.
\newblock Reduced rank regression via adaptive nuclear norm penalization.
\newblock {\em Biometrika}, 100(4):901--920, 2013.

\bibitem{velu2013multivariate}
Raja Velu and Gregory Reinsel.
\newblock {\em Multivariate reduced-rank regression: theory and applications},
  volume 136.
\newblock Springer Science \& Business Media, 2013.

\bibitem{caruana_multitask_1998}
Rich Caruana.
\newblock Multitask learning.
\newblock In {\em Learning to {Learn}}, pages 95--133. Boston, MA, 1998.

\bibitem{amit_uncovering_2007}
Yonatan Amit, Michael Fink, Nathan Srebro, and Shimon Ullman.
\newblock Uncovering shared structures in multiclass classification.
\newblock In {\em Proc. Intl. Conf. {Machine} Learning}, pages 17--24, June
  2007.

\bibitem{argyriou2008convex}
Andreas Argyriou, Theodoros Evgeniou, and Massimiliano Pontil.
\newblock Convex multi-task feature learning.
\newblock {\em Machine Learning}, 73(3):243--272, 2008.

\bibitem{ahn_nested_1988}
Sung Ahn and Gregory Reinsel.
\newblock Nested reduced-rank autoregressive models for multiple time series.
\newblock {\em Journal of the American Statistical Association},
  83(403):849--856, 1988.

\bibitem{velu_reduced_1986}
Raja Velu, Gregory Reinsel, and Dean Wichern.
\newblock Reduced rank models for multiple time series.
\newblock {\em Biometrika}, 73(1):105--118, 1986.

\bibitem{wang_forecasting_2004}
Zijun Wang and David Bessler.
\newblock Forecasting performance of multivariate time series models with full
  and reduced rank: an empirical examination.
\newblock {\em International Journal of Forecasting}, 20(4):683--695, 2004.

\bibitem{basu_low_2019}
Sumanta Basu, Xianqi Li, and George Michailidis.
\newblock Low rank and structured modeling of high-dimensional vector
  autoregressions.
\newblock {\em IEEE Transactions on Signal Processing}, 67(5):1207--1222, 2019.

\bibitem{alquier_high-dimensional_2020}
Pierre Alquier, Karine Bertin, Paul Doukhan, and Rémy Garnier.
\newblock High-dimensional {VAR} with low-rank transition.
\newblock {\em Statistics and Computing}, 30(4):1139--1153, 2020.

\bibitem{melnyk2016estimating}
Igor Melnyk and Arindam Banerjee.
\newblock Estimating structured vector autoregressive models.
\newblock In {\em Proc. Intl. Conf. Machine Learning}, pages 830--839, 2016.

\bibitem{box1977canonical}
George Box and George Tiao.
\newblock A canonical analysis of multiple time series.
\newblock {\em Biometrika}, 64(2):355--365, 1977.

\bibitem{pena_identifying_1987}
Daniel Pena and George Box.
\newblock Identifying a simplifying structure in time series.
\newblock {\em Journal of the American Statistical Association},
  82(399):836--843, 1987.

\bibitem{lam_estimation_2011}
Clifford Lam, Qiwei Yao, and Neil Bathia.
\newblock Estimation of latent factors for high-dimensional time series.
\newblock {\em Biometrika}, 98(4):901--918, 2011.

\bibitem{lam2012factor}
Clifford Lam and Qiwei Yao.
\newblock Factor modeling for high-dimensional time series: inference for the
  number of factors.
\newblock {\em The Annals of Statistics}, 40(2):694--726, 2012.

\bibitem{dong_extracting_2020}
Yining Dong, Joe Qin, and Stephen Boyd.
\newblock Extracting a low-dimensional predictable time series.
\newblock 2019.

\bibitem{stone2001blind}
James Stone.
\newblock Blind source separation using temporal predictability.
\newblock {\em Neural Computation}, 13(7):1559--1574, 2001.

\bibitem{richthofer2015predictable}
Stefan Richthofer and Laurenz Wiskott.
\newblock Predictable feature analysis.
\newblock In {\em Proc. IEEE Intl. Conf. Machine Learning and Applications},
  pages 190--196. IEEE, 2015.

\bibitem{goerg2013forecastable}
Georg Goerg.
\newblock Forecastable component analysis.
\newblock In {\em Intl. Conf. Machine Learning}, pages 64--72, 2013.

\bibitem{li2014new}
Gang Li, Joe Qin, and Donghua Zhou.
\newblock A new method of dynamic latent-variable modeling for process
  monitoring.
\newblock {\em IEEE Transactions on Industrial Electronics}, 61(11):6438--6445,
  2014.

\bibitem{zhou2016autoregressive}
Le~Zhou, Gang Li, Zhihuan Song, and Joe Qin.
\newblock Autoregressive dynamic latent variable models for process monitoring.
\newblock {\em IEEE Transactions on Control Systems Technology},
  25(1):366--373, 2016.

\bibitem{dong_dynamic_2018}
Yining Dong and Joe Qin.
\newblock Dynamic latent variable analytics for process operations and control.
\newblock {\em Computers \& Chemical Engineering}, 114:69--80, 2018.

\bibitem{dong2018novel}
Yining Dong and Joe Qin.
\newblock A novel dynamic {PCA} algorithm for dynamic data modeling and process
  monitoring.
\newblock {\em Journal of Process Control}, 67:1--11, 2018.

\bibitem{delsole2001optimally}
Timothy DelSole.
\newblock Optimally persistent patterns in time-varying fields.
\newblock {\em Journal of the Atmospheric Sciences}, 58(11):1341--1356, 2001.

\bibitem{delsole2009average1}
Timothy DelSole and Michael Tippett.
\newblock Average predictability time. {Part I}: theory.
\newblock {\em Journal of the Atmospheric Sciences}, 66(5):1172--1187, 2009.

\bibitem{delsole2009average2}
Timothy DelSole and Michael Tippett.
\newblock Average predictability time. {Part II}: Seamless diagnoses of
  predictability on multiple time scales.
\newblock {\em Journal of the Atmospheric Sciences}, 66(5):1188--1204, 2009.

\bibitem{littman_predictive_2002}
Michael Littman, Richard Sutton, and Satinder Singh.
\newblock Predictive representations of state.
\newblock {\em Proc. Adv. Neural Inf. Process. Syst.}, 14, 2002.

\bibitem{rosencrantz2004learning}
Matthew Rosencrantz, Geoff Gordon, and Sebastian Thrun.
\newblock Learning low dimensional predictive representations.
\newblock In {\em Proc. Intl. Conf. Machine Learning}, page~88, 2004.

\bibitem{boots2011closing}
Byron Boots, Sajid Siddiqi, and Geoffrey Gordon.
\newblock Closing the learning-planning loop with predictive state
  representations.
\newblock {\em The International Journal of Robotics Research}, 30(7):954--966,
  2011.

\bibitem{connor1994recurrent}
Jerome Connor, Douglas Martin, and Les Atlas.
\newblock Recurrent neural networks and robust time series prediction.
\newblock {\em IEEE Transactions on Neural Networks}, 5(2):240--254, 1994.

\bibitem{che2018recurrent}
Zhengping Che, Sanjay Purushotham, Kyunghyun Cho, David Sontag, and Yan Liu.
\newblock Recurrent neural networks for multivariate time series with missing
  values.
\newblock {\em Scientific Reports}, 8(1):1--12, 2018.

\bibitem{maggiolo2019autoregressive}
Matteo Maggiolo and Gerasimos Spanakis.
\newblock Autoregressive convolutional recurrent neural network for univariate
  and multivariate time series prediction.
\newblock {\em arXiv preprint arXiv:1903.02540}, 2019.

\bibitem{siami2018forecasting}
Sima Siami-Namini and Akbar Namin.
\newblock Forecasting economics and financial time series: {ARIMA} vs. {LSTM}.
\newblock {\em arXiv preprint arXiv:1803.06386}, 2018.

\bibitem{gers2002applying}
Felix Gers, Douglas Eck, and J{\"u}rgen Schmidhuber.
\newblock Applying {LSTM} to time series predictable through time-window
  approaches.
\newblock In {\em Neural Nets WIRN}, pages 193--200. Springer, 2002.

\bibitem{chen2015lstm}
Kai Chen, Yi~Zhou, and Fangyan Dai.
\newblock A {LSTM}-based method for stock returns prediction: A case study of
  {C}hina stock market.
\newblock In {\em Proc. IEEE Intl. Conf. Big Data}, pages 2823--2824. IEEE,
  2015.

\bibitem{livieris2020cnn}
Ioannis Livieris, Emmanuel Pintelas, and Panagiotis Pintelas.
\newblock A {CNN--LSTM} model for gold price time-series forecasting.
\newblock {\em Neural Computing and Applications}, pages 1--10, 2020.

\bibitem{afrasiabi2019dtw}
Mahlagha Afrasiabi and Muharram Mansoorizadeh.
\newblock {DTW-CNN}: time series-based human interaction prediction in videos
  using {CNN}-extracted features.
\newblock {\em The Visual Computer}, pages 1--13, 2019.

\bibitem{li2020real}
Pei Li, Mohamed Abdel-Aty, and Jinghui Yuan.
\newblock Real-time crash risk prediction on arterials based on {LSTM-CNN}.
\newblock {\em Accident Analysis \& Prevention}, 135:105371, 2020.

\bibitem{bao2017deep}
Wei Bao, Jun Yue, and Yulei Rao.
\newblock A deep learning framework for financial time series using stacked
  autoencoders and long-short term memory.
\newblock {\em PloS One}, 12(7):e0180944, 2017.

\bibitem{gensler2016deep}
Andr{\'e} Gensler, Janosch Henze, Bernhard Sick, and Nils Raabe.
\newblock Deep learning for solar power forecasting? {An} approach using
  autoencoder and {LSTM} neural networks.
\newblock In {\em Proc. IEEE Intl. Conf. Systems, Man, and Cybernetics}, pages
  2858--2865. IEEE, 2016.

\bibitem{wei2019autoencoder}
Wangyang Wei, Honghai Wu, and Huadong Ma.
\newblock An autoencoder and {LSTM}-based traffic flow prediction method.
\newblock {\em Sensors}, 19(13):2946, 2019.

\bibitem{sola1994testing}
Martin Sola and John Driffill.
\newblock Testing the term structure of interest rates using a stationary
  vector autoregression with regime switching.
\newblock {\em Journal of Economic Dynamics and Control}, 18(3-4):601--628,
  1994.

\bibitem{nyberg2018forecasting}
Henri Nyberg.
\newblock Forecasting {US} interest rates and business cycle with a nonlinear
  regime switching {VAR} model.
\newblock {\em Journal of Forecasting}, 37(1):1--15, 2018.

\bibitem{johansen1993constructing}
Tor Johansen and Bjarne Foss.
\newblock Constructing {NARMAX} models using {ARMAX} models.
\newblock {\em International Journal of Control}, 58(5):1125--1153, 1993.

\bibitem{chen1989representations}
Sheng Chen and Steve Billings.
\newblock Representations of non-linear systems: the {NARMAX} model.
\newblock {\em International Journal of Control}, 49(3):1013--1032, 1989.

\bibitem{gillard2018structured}
Jonathan Gillard and Konstantin Usevich.
\newblock Structured low-rank matrix completion for forecasting in time series
  analysis.
\newblock {\em International Journal of Forecasting}, 34(4):582--597, 2018.

\bibitem{chen2020low}
Xinyu Chen and Lijun Sun.
\newblock Low-rank autoregressive tensor completion for multivariate time
  series forecasting.
\newblock {\em arXiv preprint arXiv:2006.10436}, 2020.

\end{thebibliography}
